\documentclass[11pt]{article}

\usepackage[final]{acl}

\usepackage{times}
\usepackage{latexsym}
\usepackage{amssymb}
\usepackage{amsmath}
\usepackage[T1]{fontenc}
\usepackage{graphicx}
\usepackage{algorithm}
\usepackage{graphicx}
\usepackage{amsmath} 
\usepackage{amssymb} 
\usepackage{booktabs}
\usepackage{multirow}
\usepackage{float} 
\usepackage{amsmath}
\usepackage{newfloat}
\usepackage{listings}
\usepackage{algorithm}
\usepackage{algpseudocode}
\usepackage{xcolor}
\usepackage{amssymb}

\usepackage[utf8]{inputenc}
\usepackage[normalem]{ulem}
\useunder{\uline}{\ul}{}
\usepackage{microtype}

\usepackage{inconsolata}

\usepackage{graphicx}
\usepackage{hyperref}
\usepackage{fontawesome5}
\usepackage{simpleicons}

\definecolor{hfcolor}{HTML}{FFCC4D}
\title{Instruction Data Selection via Answer Divergence}

\author{
Bo Li\textsuperscript{\rm 1,2,3},
Mingda Wang\textsuperscript{\rm 4},
Shikun Zhang\textsuperscript{\rm 1},
Wei Ye\textsuperscript{\rm 1}\thanks{Corresponding author}
\\
\textsuperscript{\rm 1} National Engineering Research Center for Software Engineering, Peking University \\
\textsuperscript{\rm 2}School of Computer Science, Peking University \\
\textsuperscript{\rm 3}PKU-CMCC(Hubei) Joint Research Lab for LLM Industrial Applications\\
\textsuperscript{\rm 4} School of Health Sciences and Biomedical Engineering, Hebei University of Technology\\
\texttt{deepblue.lb@gmail.com, wye@pku.edu.cn}\\
\faGithub\ \href{https://github.com/WisdomShell/ADG}{WisdomShell/ADG}
\faGlobe\ \href{https://wisdomshell.github.io/ADG/}{ADG Project}
{\color{hfcolor}\simpleicon{huggingface}}\
\href{https://huggingface.co/collections/WisdomShell/adg}{HuggingFace Model}
}

\begin{document}
\maketitle
\begin{abstract}

Instruction tuning relies on large instruction–response corpora whose quality and composition strongly affect downstream performance. We propose \textbf{A}nswer \textbf{D}ivergence-\textbf{G}uided Selection \textbf{(ADG)}, which selects instruction data based on the geometric structure of multi-sample outputs. ADG draws several high-temperature generations per instruction, maps responses into an embedding space, and computes an output divergence score that jointly encodes dispersion magnitude and shape anisotropy. High scores correspond to instructions whose answers are both far apart and multi-modal, rather than clustered paraphrases along a single direction. Across two backbones  and three public instruction pools, fine-tuning on only 10K ADG-selected examples consistently outperforms strong selectors on six benchmarks spanning reasoning, knowledge, and coding. Analyses further show that both dispersion magnitude and shape anisotropy are necessary, supporting answer divergence as a practical signal for instruction data selection. Code and appendix are included in the supplementary materials.

\end{abstract}

\section{Introduction}
Large language models are typically turned into helpful assistants by instruction tuning on collections of instruction--response pairs~\cite{Ouyang2022TrainingLM,wei2022finetuned,Sanh2022MultitaskPT}.
Beyond scaling corpus size and diversity, recent work shows that performance is highly sensitive to \emph{which} examples are used under the same budget, making data selection a first-order design choice for efficient alignment~\cite{wang2023self,zhou2023lima,li2024superfiltering,he2025fine,chen2024alpagasus,Wang2024ASO,liu-etal-2025-take,Li2026DataSF,zhao2026generatingeffectivecottraces,liu2026learningcontrastssynthesizingreasoning}.

Existing selection approaches largely follow two directions.
\textbf{Data-centric} methods improve the corpus directly, for example by manual curation, filtering, rewriting, or enforcing stylistic constraints. These strategies are broadly applicable, but they do not explicitly reflect what a particular base model is missing.~\cite{zhou2023lima,li2024superfiltering,chen2024alpagasus,he2025fine,li-etal-2025-scar}~\textbf{Model-centric} methods attempt to read such information from the model itself, scoring each instruction by loss, uncertainty, learning progress, or influence, and then selecting a top-scoring subset for fine-tuning.
In practice, a common scoring design is to measure error against a \emph{single} teacher response~\cite{li2024quantity,liu2024selectit,xia2024less,Chen2025MIGAD}. However, single-reference scoring has a structural limitation in instruction tuning.
Many instructions admit multiple valid answers, and the ``reference'' response is only one realization among many.
A high loss may therefore reflect benign differences in format, tone, or reasoning style rather than genuine gaps in the model's competence. 

There is growing evidence that multi-sample behavior carries richer signals about what a model has and has not learned. Classical query-by-committee targets points with maximal disagreement across hypotheses and shows that such points concentrate near decision boundaries~\cite{Seung1992QueryBC}. Deep ensembles interpret predictive disagreement across independently trained networks as a proxy for epistemic uncertainty~\cite{Lakshminarayanan2016SimpleAS}. In LLMs, self-consistency and Tree-of-Thoughts explicitly exploit the diversity among multiple reasoning chains for the same query to obtain more reliable or stronger answers~\cite{Wang2022SelfConsistencyIC,Yao2023TreeOT}. These lines of work inspire us that data selection should also be based on how a model's outputs for the same instruction populate representation space, rather than judging an instruction by its fit to one reference.

We introduce \textbf{A}nswer \textbf{D}ivergence-\textbf{G}uided Selection \textbf{(ADG)}, a data selection framework built around this idea. ADG treats the set of responses generated at a relatively high temperature for a given instruction as a geometric object. It then computes an output divergence score that captures two complementary aspects: how widely the responses spread, and whether the spread reflects genuinely multi-directional structure rather than a near one-dimensional drift. Intuitively, high scores indicate prompts where the model's behavior is diverse in a structured way, which is more likely to reveal boundary cases than redundant supervision. To prevent a global top-$k$ from collapsing onto a few dense regions, ADG clusters instructions using semantic embeddings, and performs bin-wise selection by ranking examples within each bin according to their divergence scores.

We evaluate ADG on two backbones and three selection pools with a fixed 10K budget, and assess reasoning, knowledge, and coding on six standard benchmarks.
Across settings, ADG achieves the best or tied-best average performance and consistently outperforms strong selectors. We also conduct ablations and data-level analyses to interpret what ADG selects, including task-type shifts and a quadrant-based case study. In summary, our main contributions are as follows:

\begin{itemize}
    \item We propose ADG, a data selection method that scores instructions using multi-sample response behavior for effective instruction tuning under a fixed budget.
    \item We demonstrate consistent gains over strong selectors across two backbones, three selection pools, and six benchmarks, and provide ablations and data-level analyses that validate key design choices and characterize the selected instructions.
\end{itemize}

\begin{figure*}[h]
    \centering
    \includegraphics[width=0.89\linewidth]{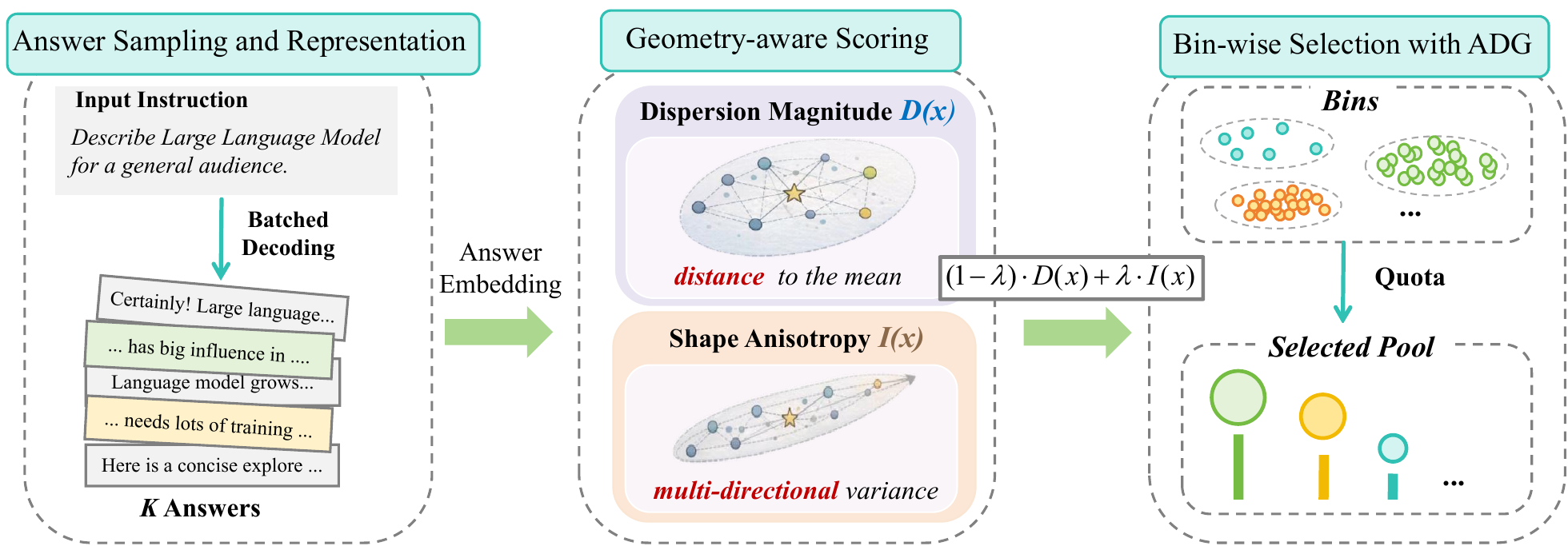}
    \caption{Overview of ADG: sample \(K\) answers per instruction, embed answers from output hidden states, score with \(D(x)\) (dispersion magnitude) and \(I(x)\) (shape anisotropy), and select top examples within semantic bins.}
    \label{fig:model}
\end{figure*}

\section{Answer Divergence-Guided Selection}

\subsection{Overview of ADG}

We aim to select the most training-effective instructions from a large pool. \textbf{A}nswer \textbf{D}ivergence-\textbf{G}uided Selection \textbf{(ADG)} scores an instruction by the model's \emph{multi-sample response behavior} under stochastic decoding. Given an instruction \(x\), we sample \(K\) answers at a relatively high temperature and embed each answer into a hidden representation, yielding a set of response vectors. Near-coincident vectors indicate near-deterministic behavior, while diverse vectors indicate competing hypotheses for resolving \(x\).

ADG summarizes this set with two complementary cues: \emph{dispersion magnitude}, measuring how widely responses spread, and \emph{shape anisotropy}, measuring whether the spread is multi-directional rather than dominated by a single direction. Instructions with both high dispersion and high anisotropy receive higher scores. Subsequent sections detail the computation from hidden states, score aggregation, and bin-wise selection for coverage.

\subsection{Answer Sampling and Representation}

For each instruction \(x\), ADG samples \(K\) stochastic answers in parallel from the base model at a temperature of \(T\):

\[
A(x)=\{a_1,\ldots,a_K\}, \qquad a_k \sim f_\theta(\cdot \mid x; T).
\]
This multi-sample protocol exposes the range of answers and reasoning modes the model currently considers plausible for the same instruction, and serves as the basis for our subsequent geometric analysis.

We represent each answer \(a_k\) using hidden states of its output tokens from the last few transformer layers \(\mathcal{L}\) (e.g., the final four layers). Let \(H_k^{(l)} \in \mathbb{R}^{T_k \times d}\) be the hidden matrix at layer \(l \in \mathcal{L}\). We compute an answer embedding by averaging over tokens and layers:
\begin{equation}
h_k=\frac{1}{|\mathcal{L}|}\sum_{l\in\mathcal{L}}\left(\frac{1}{T_k}\sum_{t=1}^{T_k}H_{k,t}^{(l)}\right),
\end{equation}
then \(L_2\)-normalize:
\begin{equation}
v_k=\frac{h_k}{\|h_k\|_2}, \quad k=1,\ldots,K.
\end{equation}
Stacking the embeddings yields
\begin{equation}
V(x)=
\begin{bmatrix}
v_1^\top\\
\vdots\\
v_K^\top
\end{bmatrix}
\in \mathbb{R}^{K \times d}.
\end{equation}

To remove the mean effect, we form the Gram matrix \(S = V(x)V(x)^\top\) and apply the centering matrix
\[
C = I_K - \frac{1}{K}\mathbf{1}_K\mathbf{1}_K^\top,
\]
where \(\mathbf{1}_K \in \mathbb{R}^K\) denotes the all-ones column vector, to remove the effect of the mean representation and obtain the centered similarity matrix
\begin{equation}
S_c = C S C = (C V(x)) (C V(x))^\top.
\end{equation}
All divergence statistics in Section~\ref{sec:adg-geometry} are computed from \(S_c\), and are therefore invariant to global translations of the answer representations.

\subsection{Geometry-aware Scoring}
\label{sec:adg-geometry}

Using the centered similarity matrix \(S_c\), ADG summarizes the multi-sample answer behavior for each instruction with two complementary statistics: overall spread and directional shape.

\paragraph{Dispersion magnitude.}
Let \(\mu = \frac{1}{K}\sum_{k=1}^K v_k\) be the mean of the unit-norm answer embeddings. We measure how far the \(K\) answers spread around their mean by
\begin{equation}
\begin{aligned}
D(x)
&= \frac{1}{K}\mathrm{tr}(S_c)
= \frac{1}{K}\sum_{k=1}^K \|v_k-\mu\|_2^2 \\
&= 1-\|\mu\|_2^2.
\end{aligned}
\end{equation}
When answers are nearly identical, \(\mu\) is close to a unit vector and \(D(x)\) is small; when answers are diverse, \(\|\mu\|_2\) decreases and \(D(x)\) increases.

\paragraph{Shape anisotropy.}
Let \(\gamma_1 \ge \gamma_2 \ge \dots \ge \gamma_K \ge 0\) be the eigenvalues of \(S_c\). We define
\begin{equation}
I(x) = 1 - \frac{\gamma_1}{\sum_{j=1}^K \gamma_j}.
\end{equation}
This is the complement of the fraction of variance explained by the leading direction: \(I(x)\) is near zero when variation is dominated by a single direction, and becomes larger when variance is distributed across multiple directions. When \(\sum_{j=1}^K \gamma_j = 0\) (equivalently \(S_c=0\)), we set \(I(x)=0\).

\paragraph{Combined ADG score.}
\(D(x)\) captures the amount of variation, while \(I(x)\) captures whether that variation is multi-directional. ADG combines them as
\begin{equation}
s(x) = (1 - \lambda) D(x) + \lambda I(x),
\end{equation}
where \(\lambda \in [0,1]\) controls the trade-off between spread and shape.

\subsection{Bin-wise Selection with ADG}
\label{sec:adg-selection}

Given a pool of instructions \(\{x_i\}_{i=1}^N\), ADG performs selection within semantic bins to avoid a single global top-\(k\) collapsing onto a few dense regions. We compute a frozen semantic embedding \(z_i=g(x_i)\in\mathbb{R}^p\) for each instruction using an encoder \(g\), then cluster \(\{z_i\}_{i=1}^N\) into \(B\) bins \(\{\mathcal{B}_1,\ldots,\mathcal{B}_B\}\) via \(k\)-means.

For each instruction \(x_i\), we compute its ADG score \(s(x_i)\). Let \(\mathcal{B}_b\) denote the index set of bin \(b\), with size \(|\mathcal{B}_b|\). Given a global budget \(M\), we allocate a bin quota
\begin{equation}
m_b = \mathrm{round}\!\left(M\cdot\frac{|\mathcal{B}_b|}{N}\right), \quad b=1,\ldots,B,
\end{equation}
and select the top \(m_b\) instructions in \(\mathcal{B}_b\) ranked by \(s(x_i)\). The final subset \(\mathcal{S}\) is the union of these bin-wise selections. This strategy preserves coverage across semantic regions while prioritizing, within each region, instructions whose sampled answers exhibit the strongest and most structurally diverse divergence.

\section{Experimental Setup and Main Results}

\begin{table*}[t]
\centering
\small
\setlength{\tabcolsep}{0.8mm}
\renewcommand{\arraystretch}{1.2}
\begin{tabular}{ccccccccccccc}
\toprule[1.5pt]
\multicolumn{1}{c|}{}                        & \multicolumn{4}{c|}{\textbf{Alpaca-GPT4}}                                               & \multicolumn{4}{c|}{\textbf{WizardLM}}                                                  & \multicolumn{4}{c}{\textbf{CoT}}                                  \\ \cline{2-13} 
\multicolumn{1}{c|}{}                        & \texttt{Reason.} & \texttt{Know.} & \texttt{Code.} & \multicolumn{1}{c|}{\textbf{Avg.Score}} & \texttt{Reason.} & \texttt{Know.} & \texttt{Code.} & \multicolumn{1}{c|}{\textbf{Avg.Score}} & \texttt{Reason.} & \texttt{Know.} & \texttt{Code.} & \textbf{Avg.Score} \\ \toprule[1.5pt]
\multicolumn{13}{l}{\textit{Backbone: LLaMA3-8B}}                                                                                                                                                                                                                                                                              \\ \toprule[1.5pt]
\multicolumn{1}{c|}{\textbf{All Data}}       & 45.95            & 57.63          & 37.87          & \multicolumn{1}{c|}{47.15}              & 51.60            & 54.20          & 42.85          & \multicolumn{1}{c|}{49.55}              & 46.18            & 47.60          & 26.88          & 40.22              \\
\multicolumn{1}{c|}{\textbf{Random}}         & 42.81            & 57.07          & 38.30          & \multicolumn{1}{c|}{46.06}              & 48.38            & 56.89          & 38.81          & \multicolumn{1}{c|}{48.03}              & 39.58            & 47.00          & 29.38          & 38.65              \\
\multicolumn{1}{c|}{\textbf{IFD}}            & 43.19            & 59.65          & 43.93          & \multicolumn{1}{c|}{48.92}              & 50.34            & 57.73          & 42.52          & \multicolumn{1}{c|}{50.19}              & 37.09            & 56.63          & 39.19          & 44.30              \\
\multicolumn{1}{c|}{\textbf{SuperFiltering}} & 46.80            & 58.46          & 42.93          & \multicolumn{1}{c|}{\uline{49.39}} & 51.09            & 57.87          & 42.11          & \multicolumn{1}{c|}{\uline{50.35}}              & 44.89            & 49.04          & 38.19          & \uline{44.03}              \\
\multicolumn{1}{c|}{\textbf{SelectIT}}       & 42.28            & 56.23          & 37.57          & \multicolumn{1}{c|}{45.36}              & 45.02            & 53.42          & 37.67          & \multicolumn{1}{c|}{45.37}              & 35.52            & 45.41          & 29.80          & 36.91              \\
\multicolumn{1}{c|}{\textbf{Rethinking}}     & 44.69            & 58.30          & 38.20          & \multicolumn{1}{c|}{47.06}              & 45.05            & 54.03          & 36.68          & \multicolumn{1}{c|}{45.25}              & 42.53            & 50.22          & 32.71          & 41.82              \\
\multicolumn{1}{c|}{\textbf{ZIP}}            & 44.50            & 56.79          & 38.90          & \multicolumn{1}{c|}{46.73}              & 44.66            & 56.70          & 36.47          & \multicolumn{1}{c|}{45.94}              & 43.60            & 51.37          & 33.64          & 42.87              \\
\multicolumn{1}{c|}{\textbf{MIG}}            &  44.46                &   57.74             &    37.99            & \multicolumn{1}{c|}{46.73}                   &     46.95             &    55.74            &    41.32            & \multicolumn{1}{c|}{48.00}                   &   42.94               &  49.63              &      35.05          &   42.54                 \\
\multicolumn{1}{c|}{\textbf{SCAR}}           &  48.10                &   59.99             &    35.87            & \multicolumn{1}{c|}{47.99}                   &   45.04               &     54.21           &     33.64           & \multicolumn{1}{c|}{44.29}                   &      40.77            &  48.75              &     29.07           &       39.53             \\
\multicolumn{1}{c|}{\textbf{ADG(Ours)}}      &   50.11         &   60.31        &      42.14     & \multicolumn{1}{c|}{\textbf{50.85}}     & 52.53            & 57.61          & 45.06          & \multicolumn{1}{c|}{\textbf{51.73}}     & 45.05            &     50.99      &    39.30       & \textbf{45.11}     \\ \toprule[1.5pt]
\multicolumn{13}{l}{\textit{Backbone: Qwen2.5-7B}}                                                                                                                                                                                                                                                                             \\\toprule[1.5pt]
\multicolumn{1}{c|}{\textbf{All Data}}       & 69.46            & 65.38          & 57.23          & \multicolumn{1}{c|}{64.02}              & 70.79            & 64.00          & 55.11          & \multicolumn{1}{c|}{63.30}              & 65.22            & 59.23          & 53.82          & 59.42              \\
\multicolumn{1}{c|}{\textbf{Random}}         & 71.69            & 66.80          & 61.80          & \multicolumn{1}{c|}{66.76}              & 72.26            & 64.00          & 57.97          & \multicolumn{1}{c|}{64.74}              & 66.34            & 58.12          & 55.94          & 60.13              \\
\multicolumn{1}{c|}{\textbf{IFD}}            & 73.54            & 66.38          & 61.79          & \multicolumn{1}{c|}{67.24}              & 73.22            & 65.99          & 53.19          & \multicolumn{1}{c|}{64.13}              & 69.19            & 59.37          & 56.95          & \uline{61.84}              \\
\multicolumn{1}{c|}{\textbf{SuperFiltering}} & 73.14            & 67.01          & 64.00          & \multicolumn{1}{c|}{\uline{68.05}}              & 73.94            & 65.23          & 57.55          & \multicolumn{1}{c|}{65.57}              & 68.43            & 59.95          & 55.14          & 61.17              \\
\multicolumn{1}{c|}{\textbf{SelectIT}}       & 71.44            & 63.83          & 61.30          & \multicolumn{1}{c|}{65.52}              & 71.79            & 64.16          & 55.17          & \multicolumn{1}{c|}{63.71}              & 66.46            & 58.89          & 54.54          & 59.96              \\
\multicolumn{1}{c|}{\textbf{Rethinking}}     & 72.93            & 66.79          & 60.78          & \multicolumn{1}{c|}{66.83}              & 72.49            & 64.59          & 55.85          & \multicolumn{1}{c|}{64.31}              & 68.30            & 59.64          & 55.42          & 61.12              \\
\multicolumn{1}{c|}{\textbf{ZIP}}            & 72.97            & 67.42          & 63.31          & \multicolumn{1}{c|}{67.90}              & 71.12            & 65.22          & 57.26          & \multicolumn{1}{c|}{64.53}              & 66.90            & 58.62          & 56.67          & 60.73              \\
\multicolumn{1}{c|}{\textbf{MIG}}            &   74.78               &    67.13            &    60.88        & \multicolumn{1}{c|}{67.59}                   &    74.10              &     65.68           &      59.28          & \multicolumn{1}{c|}{\uline{66.35}}                   &     67.90             &  59.24              &     56.03           &     61.05               \\
\multicolumn{1}{c|}{\textbf{SCAR}}           &   75.22               &    66.41            &      59.78          & \multicolumn{1}{c|}{67.10}                   &    73.65              &      64.44          &    57.36            & \multicolumn{1}{c|}{65.15}                   &    66.05              &  57.25              &      56.55          &       59.95             \\
\multicolumn{1}{c|}{\textbf{ADG(Ours)}}      & 75.43            & 67.36          & 64.40          & \multicolumn{1}{c|}{\textbf{69.06}}     & 75.07            & 66.67          & 61.19          & \multicolumn{1}{c|}{\textbf{67.64}}     & 70.55            & 59.40          & 58.17          & \textbf{62.71}     \\ \toprule[1.5pt]
\end{tabular}
\caption{
Main results with a fixed selection budget ($M{=}10$K) for selection methods across three selection datasets (Alpaca-GPT4, WizardLM, and CoT) and two backbones (LLaMA3-8B-Instruct and Qwen2.5-7B-Instruct).
\texttt{Reason.}, \texttt{Know.}, and \texttt{Code.} denote the average performance on reasoning (BBH, GSM8K), knowledge (MMLU, TruthfulQA), and code generation (HumanEval, MBPP). \textbf{Avg.Score} is computed as the unweighted average of the three capability scores.
All selection methods select 10K instances from the same pool, while All Data uses the full pool. All models are fine-tuned and evaluated with identical training and decoding configurations, and we report the mean over five random seeds to reduce randomness.}

\label{tab:main_results_avg}
\end{table*}

\subsection{Experimental Setup}
\label{sec:exp_setup}

\paragraph{Selection Pools and Budget.}
We consider three widely-used instruction corpora as selection pools: Alpaca-GPT4 (52K)~\cite{Peng2023InstructionTW}, WizardLM (70K)~\cite{xu2024wizardlm}, and CoT (100K)~\cite{wei2022chain}. Unless otherwise specified, all methods select a fixed budget of $M{=}10\text{K}$ instances from each corpus \emph{individually}. 

\paragraph{Evaluation Benchmarks and Metrics.}
We evaluate broad capabilities spanning reasoning, knowledge, and code generation. For reasoning, we use BBH~\cite{suzgun-etal-2023-challenging} and GSM8K~\cite{Cobbe2021TrainingVT} with exact match (EM). For knowledge, we use MMLU~\cite{hendrycks2021measuring} and TruthfulQA~\cite{Lin2021TruthfulQAMH} with accuracy. For coding, we use HumanEval~\cite{Chen2021EvaluatingLL} and MBPP~\cite{Austin2021ProgramSW} with pass@1. We adopt standard evaluation scripts for each benchmark, keep decoding parameters identical across methods. See Appendix~\ref{app:data} for more details.


\subsection{Backbone Models and Training Details}
\label{sec:training_details}

\paragraph{Backbones.}
We conduct experiments on LLaMA3-8B-Instruct~\cite{Dubey2024TheL3} and Qwen2.5-7B-Instruct~\cite{Yang2024Qwen25TR}. We further include results on the much smaller LLaMA3.2-1B-Instruct in the Appendix~\ref{sec:appendix_llama1b} to verify that ADG remains effective in the low-capacity regime and does not rely on large-model-specific behaviors. For each backbone, we compute the ADG score using the same backbone and then fine-tune it on the selected subset.

\paragraph{ADG Scoring and Selection Details.}
We score each instruction using the proposed ADG pipeline. Specifically, we sample $K{=}5$ candidate answers per instruction with temperature $T{=}1.4$, top-$p{=}0.9$, and max\_new\_tokens{=}180 (for ADG scoring only). These settings are used only for response sampling during ADG scoring. We obtain answer embeddings by pooling output-token hidden states and averaging the last four transformer layers. After scoring the full pool, we apply $k$-means binning with 1{,}000 clusters and select top-ranked instances within each bin under proportional quotas, finally merging all bins to form the selected subset.

\paragraph{Fine-tuning Configuration.}
We use a unified full fine-tuning recipe across all methods: 3 epochs, learning rate $1\mathrm{e}{-}5$, cosine scheduling with warmup ratio 0.03, per-device batch size 2, gradient accumulation 8, and maximum sequence length 2048. We keep the optimization and system settings (e.g., precision and distributed training) identical across methods. For every method and setting, we run fine-tuning and evaluation five times with different random seeds, and report the mean results.

\subsection{Baselines}
\label{sec:baselines}

\paragraph{Compared Methods.}
We compare ADG with a comprehensive set of strong data selection baselines, including \textbf{All Data}, \textbf{Random}, \textbf{IFD}~\cite{li2024quantity}, \textbf{SuperFiltering}~\cite{li2024superfiltering}, \textbf{SelectIT}~\cite{liu2024selectit}, \textbf{Rethinking}~\cite{Xia2024RethinkingDS}, \textbf{ZIP}~\cite{Yin2024EntropyLT}, \textbf{MIG}~\cite{Chen2025MIGAD}, and \textbf{SCAR}~\cite{li-etal-2025-scar}. For each method, we use the official open-source implementation released by the authors. We also  provide concise descriptions for all baselines in the Appendix~\ref{sec:appendix_baseline}.

\paragraph{Fairness Controls.}
We keep the fine-tuning configuration, selection budget, and evaluation protocol identical across methods; the only difference is the selected subset produced by each selector. We follow the default hyperparameters recommended by each baseline.

\subsection{Main Results}
\label{sec:main_results}

Table~\ref{tab:main_results_avg} reports results under a fixed selection budget ($M{=}10$K) across three pools and two backbones.~\footnote{The detailed results on each dataset are provided in the Appendix~\ref{app:full_results}.} ADG achieves the best average score in every setting, demonstrating that the proposed geometry-aware criterion generalizes across both model families and selection pools. Specifically, on LLaMA3-8B-Instruct, ADG improves the average score by about $2\%$--$3\%$ (relative) over the best baseline depending on the pool; on Qwen2.5-7B-Instruct, gains are typically $1\%$--$2\%$. These results indicate that ADG selects consistently more training-effective instructions under the same data budget.

The gains are most pronounced on noisier pools such as CoT: ADG surpasses All Data by +4.89 Avg.Score on LLaMA3 and +3.29 on Qwen2.5, supporting our motivation that more data is not necessarily better when supervision is redundant or unstable. Moreover, ADG yields more balanced improvements across capability groups, strengthening reasoning and coding while maintaining competitive knowledge performance, rather than trading off one axis for another.

\paragraph{Robustness to Chinese instruction data.}
We further evaluate ADG on a public Chinese instruction pool under the same setting, and observe consistent improvements over strong baselines. Full results are reported in Appendix~\ref{app:zh}.

\section{Analysis}

\subsection{Ablation Study}
\paragraph{Setup.}
All ablations are conducted on \textbf{LLaMA3-8B-Instruct} with \textbf{Alpaca-GPT4 (52K)} as the selection pool. We keep the selection budget fixed to 10K and use the same setting in the main experiments, reporting the mean over five runs.

\begin{figure}[h]
    \centering
    \includegraphics[width=0.89\linewidth]{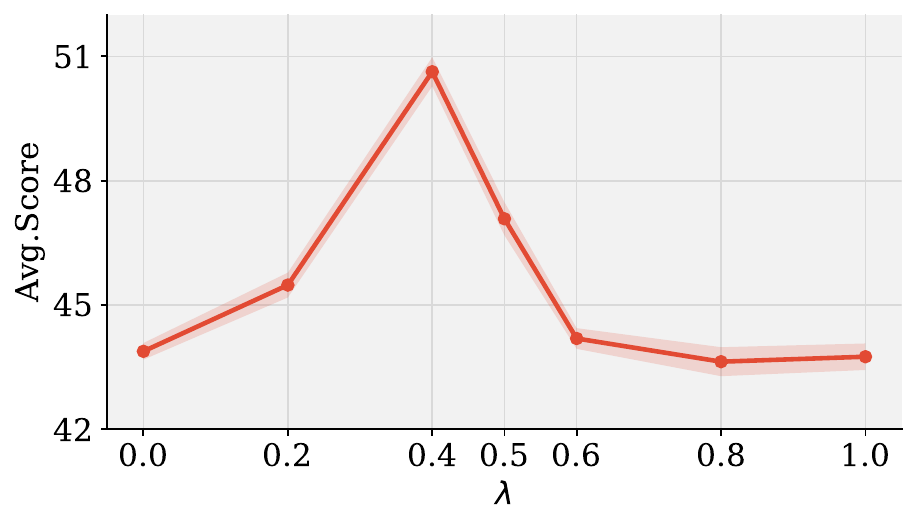}
    \caption{Effect of the fusion weight $\lambda$ on ADG. We report the mean Avg.Score over five runs, with the shaded band indicating standard deviation across five runs.}
    \label{fig:lambda_ablation}
\end{figure}

\paragraph{Ablation on the fusion weight $\lambda$.}
Figure~\ref{fig:lambda_ablation} shows a clear unimodal behavior with the best performance at $\lambda{=}0.4$. Both extremes degrade performance, indicating that the two scoring components provide complementary signals and that a moderate fusion yields the most reliable selection quality.

\begin{table}[t]
\centering
\small
\setlength{\tabcolsep}{6pt}
\begin{tabular}{l c}
\toprule[1.5pt]
\textbf{Ablation Setting} & \textbf{Avg.Score} \\
\midrule
\multicolumn{2}{l}{\textbf{(A) Score-rank}} \\
Top 10K (default)   & 50.85 \\
Middle 10K          & 42.69 \\
Tail 10K            & 37.93 \\
\midrule[1.5pt]
\multicolumn{2}{l}{\textbf{(B) Sampling temperature for scoring (top-$p{=}0.9$)}} \\
$T{=}1.0$           & 48.49 \\
$T{=}1.4$ (default) & 50.85 \\
$T{=}2.0$           & 42.92 \\
\midrule[1.5pt]
\multicolumn{2}{l}{\textbf{(C) Number of bins for stratified selection}} \\
$B{=}200$           & 45.23 \\
$B{=}500$           & 46.23 \\
$B{=}1000$ (default)& 50.85 \\
$B{=}2000$          & 45.91 \\
\midrule[1.5pt]
\multicolumn{2}{l}{\textbf{(D) Scoring components}} \\
$D(x)$ only ($\lambda{=}0$) & 43.75 \\
$I(x)$ only ($\lambda{=}1$) & 43.88 \\
\bottomrule[1.5pt]
\end{tabular}
\caption{Ablations on key components and hyperparameters of ADG: (A) selecting 10K examples from different score-rank segments, (B) the temperature used for multi-sample answer generation, (C) the number of bins ($B$) for stratified top selection, and (D) the necessity of combining the two complementary scoring components. We report mean Avg.Score over five runs.}
\label{tab:ablation_suite}
\end{table}

\paragraph{Ablation on selection ranks (Top/Middle/Tail 10K).}
Selecting the \emph{top}-ranked 10K consistently outperforms the \emph{middle} and \emph{tail} segments by a large margin. This gap confirms that ADG induces a meaningful ordering that separates training-effective instructions from lower-value supervision.

\paragraph{Ablation on sampling temperature for scoring.}
Using a moderate temperature ($T{=}1.4$) yields the best results, while lower ($T{=}1.0$) or higher ($T{=}2.0$) temperatures hurt performance. This suggests ADG benefits from a suitable level of output diversity: too low under-exposes informative variation, while too high adds noisy dispersion.

\paragraph{Ablation on the number of bins.}
The stratified selection stage exhibits a sweet spot at $B{=}1000$. Fewer bins reduce coverage and increase redundancy, whereas too many bins over-fragment the pool and weaken within-bin top selection. Overall, moderate bin granularity best balances coverage and exploitation under a fixed budget.

\paragraph{Ablation on scoring components.}
We ablate the scoring function by using only $D(x)$ or only $I(x)$, with all other settings fixed. Both variants underperform the full score, indicating that dispersion magnitude or shape anisotropy alone is insufficient. Combining $D(x)$ and $I(x)$ yields the best Avg.Score, confirming their complementarity.

\begin{figure*}[h]
    \centering
    \includegraphics[width=0.89\linewidth]{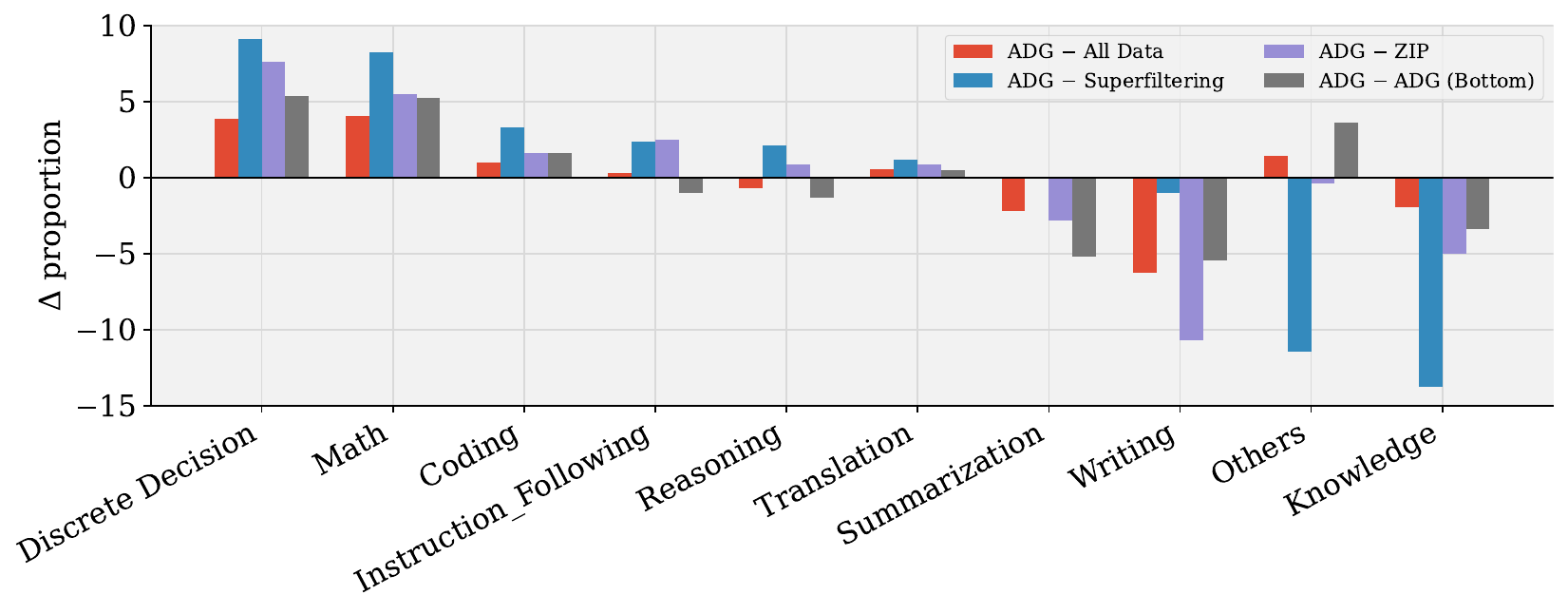}
    \caption{\textbf{Task-type composition shift of ADG-selected data.} We report $\Delta$ proportion (percentage-point change) between ADG and four reference sets (All Data, SuperFiltering, ZIP, and ADG (Bottom)) on Alpaca-GPT4. Task types are ordered by $\Delta(\text{ADG}-\text{SuperFiltering})$ from high to low.}
    \label{fig:tasktype_delta}
\end{figure*}

\subsection{Cross-backbone Generalization}
\label{sec:cross_backbone}

\paragraph{Setup.}
We study whether ADG rankings transfer across backbones when the model used for scoring differs from the model used for finetuning.
We conduct this experiment on Alpaca-GPT4 dataset with a fixed 10K selection budget.
Concretely, we first run ADG with a \emph{selection backbone} to produce a 10K subset. We then use a different \emph{fine-tune backbone} on the selected subset with the same training recipe as in our main experiments.

\begin{table}[h]
\centering
\small
\setlength{\tabcolsep}{0.8mm}
\begin{tabular}{cc|cccc}
\toprule[1.5pt]
\textbf{Fine-tune} & \textbf{Selection} & \texttt{Reason.} & \texttt{Know.} & \texttt{Code.} & \textbf{Avg.Score} \\
\toprule[1.5pt]
Qwen2.5 & Qwen2.5 & 75.43 & 67.36 & 64.40 & 69.06 \\
Qwen2.5 & LLaMA3  & 76.09 & 67.11 & 62.69 & 68.63 \\
LLaMA3  & LLaMA3  & 50.11 & 60.31 & 42.14 & 50.85 \\
LLaMA3  & Qwen2.5 & 49.84 & 59.61 & 41.11 & 50.19 \\
\toprule[1.5pt]
\end{tabular}
\caption{Cross-backbone generalization of ADG on Alpaca-GPT4 dataset with a 10K budget: we score and select data using the specified selection backbone, then fine-tune the specified backbone.
}
\label{tab:cross_backbone_full}
\end{table}

\paragraph{Results.}
Table~\ref{tab:cross_backbone_full} reports the cross-backbone transfer matrix between Qwen2.5 and LLaMA3. Overall, using a different backbone for scoring and selection results in only a small drop in Avg.Score, suggesting that ADG captures a largely transferable ranking signal rather than one that is tightly coupled to a specific backbone. Notably, the transfer gap is more pronounced on coding than on reasoning/knowledge. A plausible explanation is that coding instructions are more sensitive to backbone-specific inductive biases about program structure and formatting; consequently, divergence geometry estimated by a different backbone may place relatively more weight on stylistic variability and less on correctness-relevant signals, leading to a slightly larger mismatch in the selected subset.

\subsection{Selection Efficiency}
\label{sec:efficiency}

\paragraph{Setup}
We report the end-to-end selection time on Alpaca-GPT4 (\(N{=}52{,}002\)) for constructing a 10K subset.
All methods are run on the same machine with 4 NVIDIA A800 GPUs in a consistent runtime environment. The reported wall-clock time covers the full pipeline, including data loading, scoring (and answer sampling/generation when applicable), clustering/binning, and ranking, thus reflecting both GPU and CPU stages.

\begin{table}[h]
\centering
\small
\setlength{\tabcolsep}{3.5mm}
\begin{tabular}{lcc}
\toprule[1.5pt]
\textbf{Selector} & Time (h)$\downarrow$ & Ex/s$\uparrow$ \\
\midrule
SuperFiltering & 1.05 & 13.76 \\
ADG (ours)     & 1.78 & 8.10  \\
ZIP            & 3.17 & 4.56  \\
MIG            & 3.92 & 3.69  \\
\bottomrule[1.5pt]
\end{tabular}
\caption{End-to-end selection efficiency on Alpaca-GPT4 (\(N{=}52{,}002\)) for producing a 10K subset using 4 GPUs. Ex/s is computed as \(N/(\text{time}\times 3600)\).}
\label{tab:efficiency}
\end{table}

\paragraph{Results.}
Table~\ref{tab:efficiency} shows that ADG incurs a moderate selection cost.
ADG completes end-to-end selection in 1.78 hours (8.10 ex/s).
The overhead mainly comes from multi-sample decoding and representation extraction, which are necessary for computing divergence-based scores. In practice, selection is an offline, one-time preprocessing step, and ADG offers a flexible cost-quality trade-off via \(K\) and the maximum generation length (see Appendix~\ref{app:eff_tradeoff}).

\subsection{Robustness to Alternative Output Representation Choices}
\label{app:layer_window_ablation}

\textbf{Setup.} Our default ADG setting represents each sampled response by averaging the output hidden states from the last four transformer layers. We use this same configuration across all selection pools and backbones, without any per-dataset tuning. This choice is also consistent with recent findings that middle-to-late layers of LLMs often provide stronger and more stable representations for downstream adaptation and transfer than early layers \citep{Raja2025EvaluatingGA,Skean2025LayerBL}. To examine the robustness of ADG to alternative representation choices, we shift the 4-layer embedding window across the network. We use LLaMA3-8B-Instruct as the backbone and Alpaca-GPT4 as the instruction pool, with the same 10K selection budget as in the main experiments. All other settings are kept identical to the default ADG configuration.

\begin{table}[t]
\centering
\small
\setlength{\tabcolsep}{6pt}
\begin{tabular}{lcccc}
\toprule[1.5pt]
Layer & \texttt{Reason.} & \texttt{Know.} & \texttt{Code.} & Avg.Score \\
\midrule[1.5pt]
0--3   & 44.46 & 57.52 & 36.87 & 46.28 \\
4--7   & 42.81 & 57.54 & 36.59 & 45.65 \\
8--11  & 38.37 & 57.11 & 35.97 & 43.82 \\
12--15 & 42.74 & 56.92 & 38.08 & 45.91 \\
16--19 & 47.62 & 59.27 & 40.41 & 49.10 \\
20--23 & 50.06 & 54.82 & 39.80 & 48.23 \\
24--27 & 48.52 & 57.96 & \textbf{42.63} & 49.70 \\
28--31 (ours) & \textbf{50.11} & \textbf{60.31} & 42.14 & \textbf{50.85} \\
\bottomrule[1.5pt]
\end{tabular}
\caption{Robustness of ADG to different 4-layer output representation windows on LLaMA3-8B-Instruct with Alpaca-GPT4 under a fixed 10K selection budget. Reason., Know., and Code. denote the average performance on reasoning (BBH, GSM8K), knowledge (MMLU, TruthfulQA), and code generation (HumanEval, MBPP), respectively. Avg.Score is the unweighted average of the three capability scores.}
\label{tab:layer_window_ablation}
\end{table}

\textbf{Results.} Table~\ref{tab:layer_window_ablation} shows that the best performance is achieved by the last-four-layer window (28--31), which reaches an Avg.Score of 50.85. Nearby higher-layer choices remain competitive, such as 24--27 (49.70) and 16--19 (49.10), whereas early-layer windows perform clearly worse, such as 0--3 (46.28) and 8--11 (43.82). These results suggest that ADG is not overly sensitive to the exact layer choice as long as higher-layer representations are used, and that averaging the last four layers provides a simple and robust default.

\subsection{Task-type Distribution of Selected Instructions}
\label{sec:analysis_tasktype}

\paragraph{Setup.}
We categorize Alpaca-GPT4 instructions into ten coarse task types using GPT-4o as a classifier, and analyze how different selectors change the \emph{composition} of the resulting 10K subset.
Figure~\ref{fig:tasktype_delta} reports $\Delta$ proportion (percentage-point change) between the ADG-selected subset and four reference sets (All Data, SuperFiltering, ZIP, and ADG Bottom), with task types ordered by $\Delta(\text{ADG}-\text{SuperFiltering})$ from high to low. The GPT-4o classification prompt is provided in the Appendix~\ref{app:prompt}.

\paragraph{ADG rebalances the selected subset toward structured tasks.} As shown in Figure~\ref{fig:tasktype_delta}, ADG consistently increases the share of Discrete Decision and Math (with a smaller gain on Coding), while mainly reducing writing and Knowledge-heavy prompts.
The same pattern holds when comparing against the full pool and the ADG-bottom subset, suggesting that the shift is robust to the choice of reference.


\paragraph{Why Discrete Decision and Math are favored.}
This trend aligns with ADG's multi-sample divergence scoring: Discrete Decision tasks often yield a few well-separated output modes corresponding to competing choices, and Math tasks amplify small reasoning differences into different final answers.
In both cases, higher divergence more often reflects verifiable disagreement than surface-level paraphrasing, making these instructions more boundary-revealing and learnable.


Overall, ADG tends to prioritize regimes where response disagreement more often corresponds to \emph{learnable decision boundaries} and \emph{verifiable reasoning}, providing a data-level perspective on its stronger downstream generalization. Appendix~\ref{app:sample_property} further provides an LLM-based sample-property analysis along five dimensions, offering complementary evidence about the characteristics of ADG-selected data.

\begin{figure*}[h]
    \centering
    \includegraphics[width=0.99\linewidth]{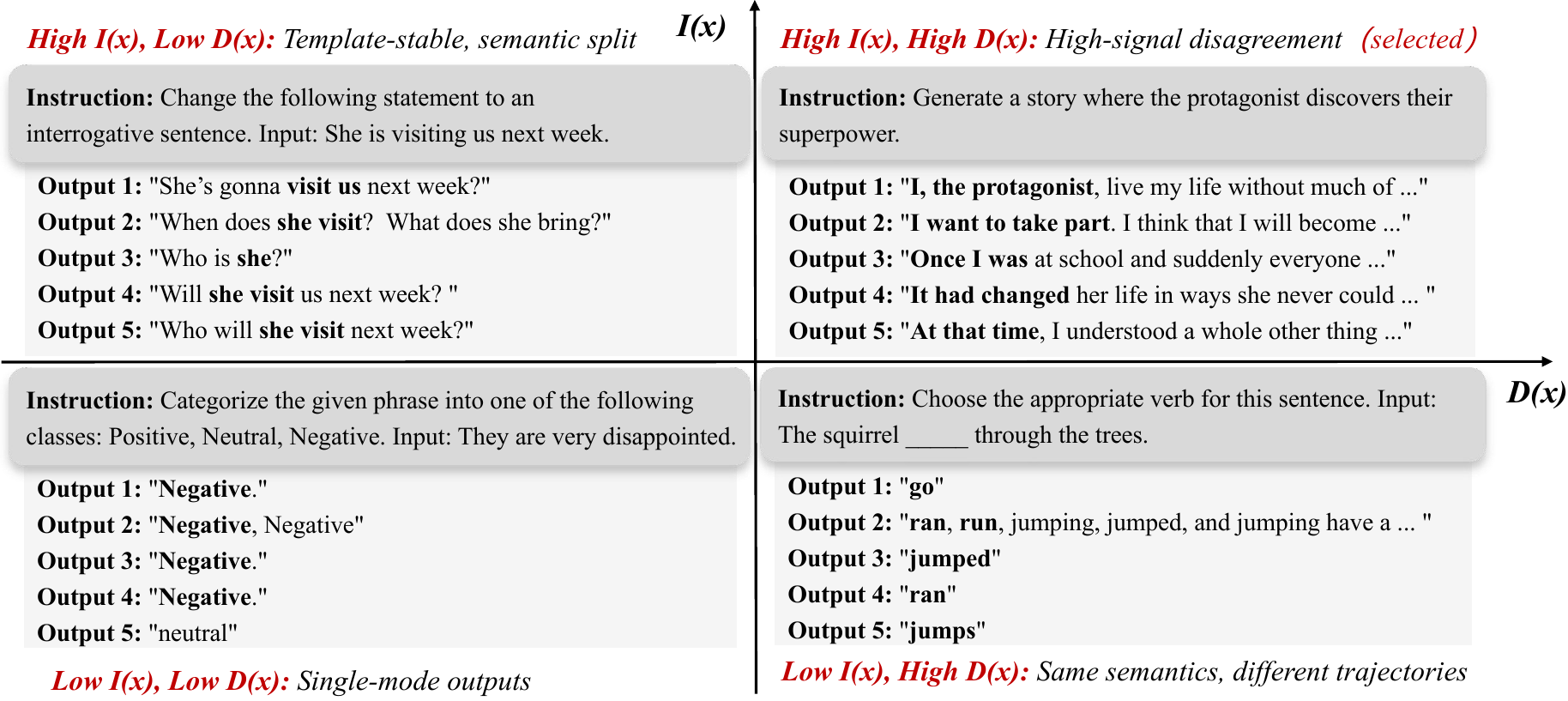}
    \caption{A quadrant-based case study of ADG using $D(x)$ (Dispersion magnitude) and $I(x)$ (Shape anisotropy). Each quadrant shows one instruction with five sampled outputs, illustrating distinct disagreement patterns; ADG favors the high-$D(x)$ and high-$I(x)$ region (top-right).}
    \label{fig:case}
\end{figure*}

\subsection{Quadrant-based Case Study}
\label{sec:case_quadrant}

\paragraph{Setup.} We interpret ADG by mapping each instruction onto a 2D plane of $D(x)$ (Dispersion magnitude) and $I(x)$ (Shape anisotropy).
For each instruction, we sample five responses and illustrate the disagreement patterns; Figure~\ref{fig:case} shows one representative example per quadrant.
The four regions correspond to distinct types of behavioral variation, explaining why ADG favors the high-$D(x)$ and high-$I(x)$ regime.

\paragraph{Low $I(x)$, Low $D(x)$: single-mode, low-signal.}
Outputs nearly collapse to one answer (often with minor formatting noise), so sampling reveals little meaningful disagreement and mainly reinforces already-stable behavior. 


\paragraph{High $I(x)$, Low $D(x)$: template-stable with a localized split.}
Responses share a consistent structure, but differ on a specific semantic pivot (e.g., who/when/what), yielding a pronounced shape (high $I(x)$) yet limited spread.
Such instances often indicate a localized decision boundary: the template is learned, while the key choice remains uncertain.


\paragraph{Low $I(x)$, High $D(x)$: diffuse spread from style/trajectory.}
Responses vary widely in verbosity or reasoning style (short vs.\ long, different phrasing), producing high dispersion without a clear anisotropic structure.
This shows that $D(x)$ alone can be dominated by non-semantic variation, and $I(x)$ helps separate structured disagreement from expression-level drift.


\paragraph{High $I(x)$, High $D(x)$: high-signal disagreement (selected).}
Samples are both widely spread and clearly structured, suggesting competing semantic routes rather than mere paraphrases.
These instances are boundary-revealing and provide higher supervision gain under a fixed budget by pushing the model toward consistent decisions across genuinely multi-modal solution spaces.


Overall, neither high dispersion nor high anisotropy alone is sufficient. ADG prioritizes instances with \emph{both} high $D(x)$ and high $I(x)$, where disagreement is strong yet structured, making it more likely to correspond to learnable decision boundaries instead of noisy stylistic variability.

\section{Related Work}

Early instruction tuning showed that curated instruction--response pairs can substantially improve alignment and generalization, motivating \emph{data-centric} pipelines that enhance corpus quality via manual or LLM-assisted cleaning, rewriting, and diversity control~\cite{wang2023self,zhou2023lima,chen2024alpagasus,li2024superfiltering,he2025fine,li2024quantity,zhou2026fincardscardbasedanalystreranking,Li2026DataSF}. In parallel, \emph{model-centric} selectors score examples using a reference model (e.g., loss/uncertainty/influence) and then select top-ranked subsets, with recent work also analyzing their behavior and biases~\cite{liu2024selectit,xia2024less,Mekala2024SmallerLM,liu-etal-2025-take,Wang2024ASO,Xia2024RethinkingDS}. Beyond per-example scoring, several methods incorporate global information structure or stylistic signals into selection~\cite{Chen2025MIGAD,li-etal-2025-scar}. Despite these advances, most methods still reduce each example to a single scalar from a single forward pass and rely on global top-$k$ selection, making them sensitive to format, domain, and length. ADG adapts multi-sample disagreement ideas to data selection~\cite{Seung1992QueryBC,Lakshminarayanan2016SimpleAS,Wang2022SelfConsistencyIC,Yao2023TreeOT}. It scores each \emph{instruction} by the geometric structure of its multi-sample outputs in an embedding space, using dispersion magnitude and shape anisotropy with selection with proportional quotas, providing a distribution-aware signal complementary to scalar loss/uncertainty and style-based selectors.

\section{Conclusion}
We introduced ADG, a geometry-based, model-centric data selection method that leverages multi-sample behavior to identify training-effective instructions. By jointly modeling dispersion magnitude and shape anisotropy and selecting within semantic bins, ADG prioritizes instances where disagreement is both strong and structured, while avoiding collapse onto length- or domain-biased regions. Experiments across two backbones and three instruction pools show consistent gains with only 10K selected examples, and analyses further validate the complementarity of the two geometric cues. These results suggest answer divergence is a simple, scalable, and broadly applicable organizing principle for efficient instruction data selection.

\section*{Limitations}
First, ADG uses a fixed design for answer-embedding extraction (mean-pooling output-token hidden states and averaging late layers). While alternative pooling and layer choices may further improve the signal-to-noise trade-off, we follow common settings in prior work and validate key choices with targeted ablations.
Second, ADG uses multi-sample disagreement structure as an indirect proxy for training value. In some cases, high divergence may also stem from prompt ambiguity or underspecified supervision. Combining ADG with lightweight quality or clarity filters is a promising direction for future work.

\bibliography{custom}

\clearpage
\appendix
\section*{Appendix}
\section{Dataset and Evaluation Protocol}\label{app:data}

\begin{table}[h]
\begin{tabular}{cc}
\bottomrule[1.5pt]
\multicolumn{1}{l|}{\textbf{Dataset}}        & \textbf{Number Instances} \\ \bottomrule[1.5pt]
\multicolumn{2}{l}{\textit{Training}}                                    \\ \hline
\multicolumn{1}{l|}{\textbf{Alpaca-GPT4}}    & 52,002                    \\
\multicolumn{1}{l|}{\textbf{WizardLM}}       & 70,000                    \\
\multicolumn{1}{l|}{\textbf{CoT}}            & 100,000                   \\ 
\multicolumn{1}{l|}{\textbf{Alpaca-GPT4-ZH}} & 42,677                    \\ \hline
\multicolumn{2}{l}{\textit{Evaluation}}                                  \\ \hline
\multicolumn{1}{l|}{\textbf{BBH}}            & 6,511                     \\
\multicolumn{1}{l|}{\textbf{GSM8K}}          & 1,319                     \\
\multicolumn{1}{l|}{\textbf{HumanEval}}      & 164                       \\
\multicolumn{1}{l|}{\textbf{MBPP}}           & 500                       \\
\multicolumn{1}{l|}{\textbf{MMLU}}           & 56,168                    \\
\multicolumn{1}{l|}{\textbf{TruthfulQA}}     & 5,882                     \\ 
\multicolumn{1}{l|}{\textbf{CMMLU(ZH)}}      & 23,232                    \\
\multicolumn{1}{l|}{\textbf{C-Eval(ZH)}}     & 2,764                     \\ \bottomrule[1.5pt]
\end{tabular}
\caption{Datasets used in our experiments. We report the training selection pools and evaluation benchmarks, together with the number of instances for each dataset.}
\label{datasets}
\end{table}

\paragraph{Dataset.} We evaluate ADG on three English instruction-tuning pools (Alpaca-GPT4, WizardLM, and CoT) and additionally include a public Chinese instruction pool (Alpaca-GPT4-ZH) for robustness under language shift. 
For evaluation, we cover a broad set of widely used benchmarks spanning reasoning (BBH, GSM8K), coding (HumanEval, MBPP), and knowledge (MMLU, TruthfulQA), and further report results on two Chinese benchmarks (CMMLU and C-Eval). 
Table~\ref{datasets} lists the dataset sizes used throughout the paper.

\paragraph{Evaluation Protocol.} We evaluate all fine-tuned models using the open-source \textit{EleutherAI Language Model Evaluation Harness} (\texttt{lm-evaluation-harness}).\footnote{\url{https://github.com/EleutherAI/lm-evaluation-harness}}
To ensure fair comparison across selection methods, we keep the evaluation pipeline fixed and use the harness' default task configurations for all.
In particular, we adopt the official few-shot settings for each benchmark:
BBH (3-shot), GSM8K (8-shot), TruthfulQA-MC2 (0-shot), MBPP (1-shot), HumanEval (0-shot), and MMLU (4-shot).
We use the harness' default decoding and scoring options for each task.
All reported results are obtained under this unified evaluation setup.

\section{Task-Type Classification Prompt}\label{app:prompt}

\begin{figure}[h]
    \centering
    \includegraphics[width=0.99\linewidth]{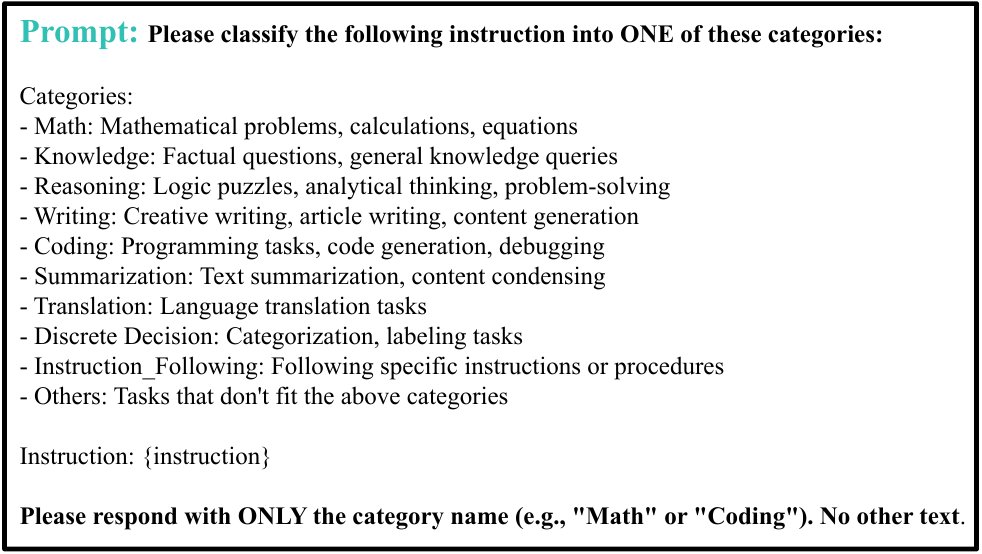}
    \caption{Task-Type classification prompt used in this paper.}
    \label{fig:task_type}
\end{figure}

To analyze task-type distribution shifts (Section~\ref{sec:analysis_tasktype}), we use GPT-4o to assign each Alpaca-GPT4 instruction to one of ten coarse categories (e.g., Math, Coding, and Others). 
Figure~\ref{fig:task_type} shows the exact prompt template. 
We require the model to output \emph{only} the category name to reduce formatting variance and ease automatic parsing.

\section{Robustness to Chinese Instructions}
\label{app:zh}

\begin{table}[h]
\centering
\small
\setlength{\tabcolsep}{7pt}
\begin{tabular}{lccc}
\toprule[1.5pt]
\textbf{Selector} &
\textbf{Avg.Score} &
\textbf{CMMLU}&
\textbf{C-Eval}\\
\midrule[1.5pt]
All Data        & 40.84 & 41.11 & 40.56 \\
Random          & 41.32 & 42.08 & 40.56 \\
SuperFiltering  & 44.50 & 44.02 & 44.97 \\
Rethinking      & 42.40 & 42.53 & 42.27 \\
\midrule
\textbf{ADG (Ours)} & \textbf{45.54} & \textbf{45.09} & \textbf{45.99} \\
\bottomrule[1.5pt]
\end{tabular}
\caption{Results on Chinese benchmarks. We select 10K instances from Alpaca-GPT4-ZH, fine-tune LLaMA3-8B-Instruct and evaluate on CMMLU and C-Eval. We report the mean over five runs.}
\label{tab:crosslingual_zh}
\end{table}

\paragraph{Setup.}
To assess robustness beyond English, we additionally evaluate on Alpaca-GPT4-ZH, a publicly available Chinese instruction-tuning pool with 42.6K instances~\cite{Peng2023InstructionTW}. We then select 10K samples using each selector, fine-tune LLaMA3-8B-Instruct under the same recipe as the main experiments, and evaluate the resulting models on two Chinese knowledge benchmarks, CMMLU~\cite{Li2023CMMLUMM} and C-Eval~\cite{huang2023ceval}, and we report the average score over five runs.

\paragraph{Results and discussion.}
As shown in Table~\ref{tab:crosslingual_zh}, ADG achieves the best performance on the aggregated score and on both Chinese benchmarks, demonstrating that our selection criterion transfers well beyond the original English evaluation suite. Compared with the strongest baseline selector (SuperFiltering), ADG improves the overall Avg.Score by about \textbf{+2.3\% relative}, with consistent gains on CMMLU (\textbf{+2.4\%}) and C-Eval (\textbf{+2.3\%}). Notably, training on the Chinese instruction-tuning pool is not always optimal, and even random selection can be competitive, suggesting that the Chinese instruction pool still contains redundancy and noisy supervision. These results suggest that ADG captures model-behavior signals that remain informative when the instruction language changes, rather than relying on language-specific heuristics.

\section{Full Benchmark Results}
\label{app:full_results}

To provide a complete view beyond the aggregated averages reported in the main text, we present detailed results for \textbf{both backbones} on \textbf{all six evaluation benchmarks} in this appendix.
Specifically, we report per-benchmark scores for models fine-tuned on the 10K subsets selected by each method in Table~\ref{app:llama} and Table~\ref{app:qwen}, using the same fine-tuning budget and evaluation protocol as in Section~\ref{sec:main_results}.
These tables clarify where ADG yields the largest gains and help verify that improvements are not driven by a single benchmark, but are consistently observed across reasoning, knowledge, and coding evaluations.

\begin{table*}[h]
\centering
\setlength{\tabcolsep}{6pt}
\begin{tabular}{lccccc}
\bottomrule[1.5pt]
\textbf{Method} & \textbf{Cognitive} & \textbf{Instruction} & \textbf{Response} & \textbf{Educational} & \textbf{Scalability} \\
 & \textbf{Complexity} & \textbf{Precision} & \textbf{Quality} & \textbf{Value} & \textbf{Potential} \\
\bottomrule[1.5pt]
Random          & 3.94 & 8.10 & 7.64 & 5.00 & 5.56 \\
SuperFiltering  & 5.79 & 8.56 & 8.07 & 6.83 & 7.27 \\
ADG\_Bottom     & 3.02 & 8.02 & 7.74 & 4.13 & 4.80 \\
ADG\_Top(default)        & \textbf{6.37} & \textbf{8.64} & \textbf{8.16} & \textbf{7.11} & \textbf{7.48} \\
\bottomrule[1.5pt]
\end{tabular}
\caption{LLM-based rating scores (GPT-4o) on five sample-quality dimensions. Each value is the mean over 1,000 randomly sampled instruction--response pairs from the corresponding subset.}
\label{tab:sample_property}
\end{table*}

\begin{table*}[h]
\centering
\begin{tabular}{c c c c c c c c}
\toprule[1.5pt]
$K$ & Avg.Score & BBH & GSM8K & MMLU & TruthfulQA & HumanEval & MBPP \\
\midrule[1.5pt]
3  & 49.32 & 53.05 & 42.00 & 62.40 & 54.32 & 37.80 & \textbf{46.40} \\
5 (default)  & \textbf{50.85} & \textbf{57.38} & \textbf{42.84} & \textbf{62.67} & \textbf{57.94} & \textbf{42.07} & 42.20 \\
7  & 46.79 & 54.31 & 34.87 & 61.24 & 54.77 & 34.76 & 40.80 \\
10 & 45.72 & 50.38 & 37.15 & 59.93 & 55.73 & 33.54 & 37.60 \\
\bottomrule[1.5pt]
\end{tabular}
\caption{Ablation on the number of sampled answers $K$ for ADG scoring and selection, evaluated with a LLaMA3-8B-Instruct backbone on the Alpaca-GPT4 pool. Bold indicates the best score per column.}
\label{tab:ablation_k}
\end{table*}

\begin{figure}[h]
    \centering
    \includegraphics[width=0.99\linewidth]{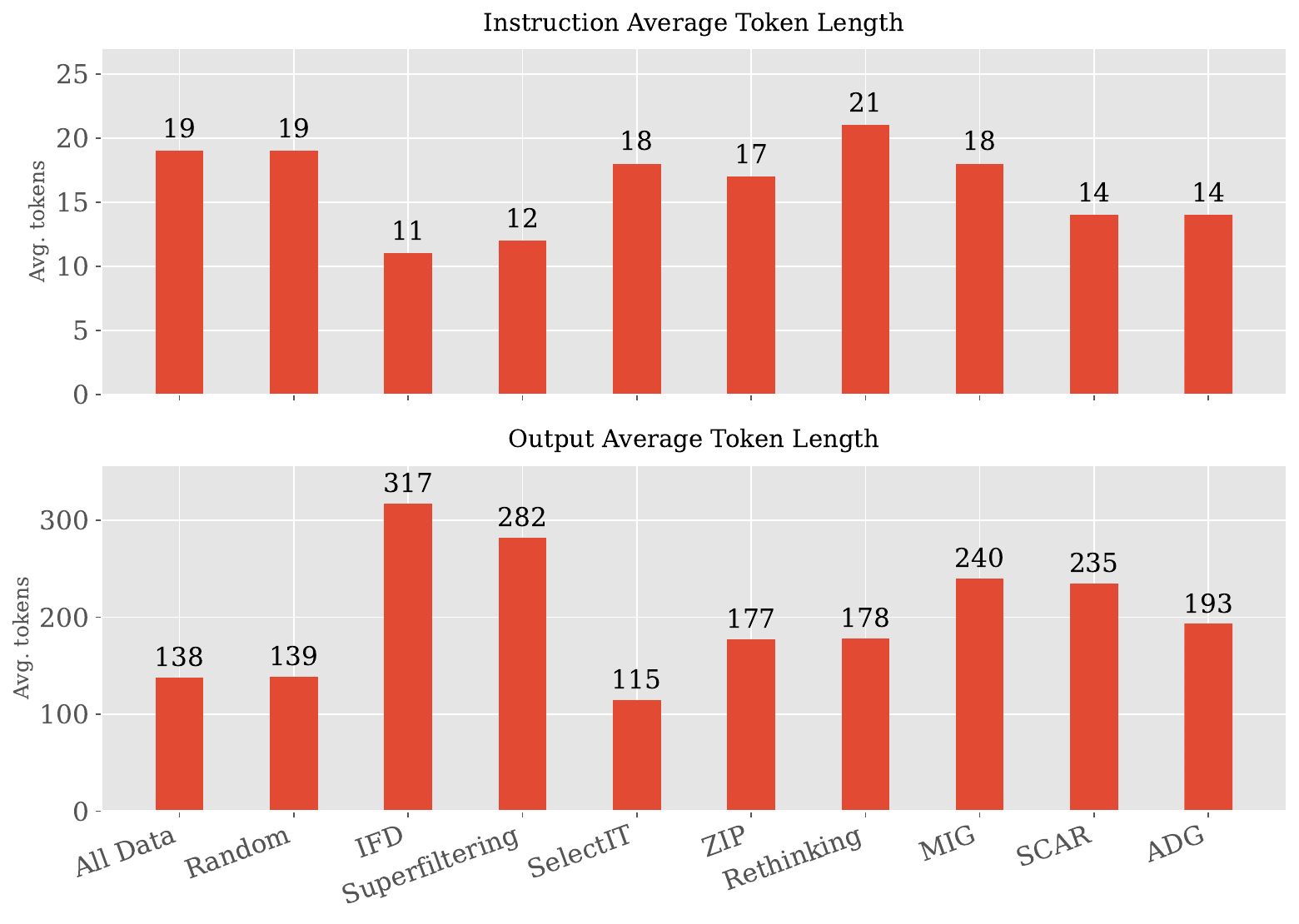}
    \caption{Average token length of the selected subsets, reporting instruction (input) and response (output) lengths under the same tokenizer.}
    \label{fig:length_stats}
\end{figure}

\section{Instruction and Response Length Statistics}
\label{sec:analysis_length}

Figure~\ref{fig:length_stats} summarizes the \emph{average token length} of the selected subsets, measured separately for the instruction (input) and the model-generated response (output) using the same tokenizer. Overall, ADG produces \textbf{moderate-length} training data: its selected instructions are shorter than the full pool (14 vs.\ 19 tokens on average), while its responses are \emph{not} excessively long (193 tokens), staying close to ZIP/Rethinking (177--178) and far below the very long-response subsets (e.g., IFD at 317 and SuperFiltering at 282). In contrast, some baselines exhibit clearer length skew, such as selecting very short instructions (IFD, SuperFiltering) or very short outputs (SelectIT at 115). These patterns suggest that ADG's improvements are unlikely to come from simply shifting toward longer (or shorter) generations; instead, ADG selects samples with more informative multi-answer behavior while keeping response verbosity within a reasonable range.

\begin{table}[h]
\centering
\small
\setlength{\tabcolsep}{2pt}
\begin{tabular}{l l c c c c}
\toprule
 $M$ & Method & BBH & MMLU & HumEval & Avg.Score \\
\midrule
3K  & Random         & 51.37 & 61.26 & 34.76 & 49.13 \\
    & SuperFiltering & 51.94 & 61.65 & 42.68 & 52.09 \\
    & ADG            & \textbf{52.46} & \textbf{62.92} & \textbf{43.29} & \textbf{52.89} \\
\midrule
5K  & Random         & 51.26 & \textbf{61.72} & 35.80 & 49.59 \\
    & SuperFiltering & \textbf{54.04} & 59.97 & 37.80 & 50.60 \\
    & ADG            & 53.68 & 61.06 & \textbf{43.34} & \textbf{52.69} \\
\midrule
10K & Random         & 52.35 & 60.48 & 37.80 & 50.21 \\
    & SuperFiltering & 53.86 & 60.13 & 40.85 & 51.61 \\
    & ADG            & \textbf{57.38} & \textbf{62.67} & \textbf{42.07} & \textbf{54.04} \\
\midrule
20K & Random         & 53.37 & 61.54 & \textbf{36.59} & 50.50 \\
    & SuperFiltering & 55.07 & 61.12 & 35.37 & 50.52 \\
    & ADG            & \textbf{56.44} & \textbf{62.43} & 35.98 & \textbf{51.62} \\
\midrule
25K & Random         & 54.58 & 61.71 & 31.71 & 49.33 \\
    & SuperFiltering & 51.97 & \textbf{61.85} & 42.73 & 52.18 \\
    & ADG            & \textbf{55.66} & 61.37 & \textbf{44.98} & \textbf{54.00} \\
\bottomrule
\end{tabular}
\caption{Budget-scaling results on LLaMA3-8B-Instruct with Alpaca-GPT4 as the selection pool. Avg.Score is the unweighted average of the three representative benchmarks.}
\label{tab:budget_scaling}
\end{table}

\section{Behavior under Different Selection Budgets}
\label{app:budget_scaling}

\textbf{Setup.} We study how ADG behaves as the selection budget increases. Due to rebuttal-time constraints, we conduct a focused budget-scaling experiment on LLaMA3-8B-Instruct using Alpaca-GPT4 (52K) as the selection pool. We evaluate one representative benchmark for each capability group: BBH for reasoning, MMLU for knowledge, and HumanEval for coding. We vary the selection budget over $M \in \{3\text{K}, 5\text{K}, 10\text{K}, 20\text{K}, 25\text{K}\}$, corresponding to approximately $6\%$, $10\%$, $20\%$, $39\%$, and $48\%$ of the full pool, respectively. All methods use the same SFT protocol, and we compare ADG against Random and SuperFiltering under identical settings.

\textbf{Results.} Table~\ref{tab:budget_scaling} shows that ADG consistently outperforms Random across all selection budgets, and also surpasses SuperFiltering in Avg.Score at every budget we test. The gains are already visible under small budgets, where ADG achieves 52.89 at 3K and 52.69 at 5K, and remain strong at larger budgets, including 54.04 at 10K and 54.00 at 25K. At the same time, the gains do not increase monotonically with the budget. In particular, performance improves from 3K to 10K, but does not continue to rise thereafter, suggesting that the benefit of simply increasing the selected subset gradually saturates. This trend is consistent with our overall motivation: under a fixed training recipe, selection quality matters more than selecting more data indiscriminately.

\section{Sample Property Analysis via LLM-based Rating}
\label{app:sample_property}

\paragraph{Goal and Evaluation protocol.}
We analyze whether examples selected by ADG exhibit known properties of high-quality instruction-tuning data, beyond downstream benchmark scores. In particular, we test if ADG tends to surface instructions that are (i) cognitively challenging yet learnable, (ii) well-specified, and (iii) paired with strong responses that are useful for generalization.

We use GPT-4o as an automatic rater and apply a uniform sampling protocol across all compared subsets.
For each subset produced by a selection method (e.g., Random, SuperFiltering, ADG\_Top, and ADG\_Bottom), we \textbf{randomly sample 1,000 examples} and ask GPT-4o to assign \emph{pointwise} scores to each example along the five dimensions defined below.
We then report the \textbf{mean rating} over the 1,000 scored examples for each subset.
This design keeps the evaluation cost comparable across methods and avoids bias from manually selecting only the best-looking samples.

\paragraph{Rating dimensions.}
Each instruction--response pair is rated on a 1--10 scale over five aspects:
\begin{itemize}
    \item \textbf{Cognitive complexity}, reflecting the degree of multi-step reasoning, abstraction, or domain knowledge required;
    \item \textbf{Instruction precision}, measuring clarity, specificity, and operationality of the instruction;
    \item  \textbf{Response quality}, measuring correctness, completeness, and directness of the response;
    \item \textbf{Educational value}, assessing whether the pair teaches useful reasoning patterns or domain knowledge;
    \item \textbf{Scalability potential}, estimating how broadly the learned pattern can transfer to similar tasks.
\end{itemize}

\paragraph{Findings.}
Table~\ref{tab:sample_property} shows that ADG-selected examples score consistently higher than baselines across all five dimensions.
Compared with Random and the ADG\_Bottom subset, ADG\_Top exhibits notably higher \emph{cognitive complexity} and \emph{educational value}, suggesting that ADG tends to prioritize training signals that are both challenging and instructive rather than trivial or low-yield.
ADG\_Top also maintains strong \emph{instruction precision} and \emph{response quality}, indicating that the higher complexity does not come from ambiguous prompts or poorly formed supervision.
Overall, this analysis provides an interpretable, data-level view of \emph{why} ADG-selected data leads to better downstream performance: it preferentially surfaces examples that are clearer, more instructive, and more transferable.

\section{Effect of the Number of Sampled Answers $K$}
\label{sec:ablation_k}

We ablate the number of sampled answers $K$, which controls how many outputs we generate per instruction to estimate the divergence-based score.
We run this study on the Alpaca-GPT4 pool using LLaMA3-8B-Instruct as the backbone. All other settings follow the best configuration reported in the main paper, including the same decoding setup for answer sampling, the same feature extraction, the same binning strategy, and the same selection budget.

\begin{figure}[h]
    \centering
    \includegraphics[width=0.99\linewidth]{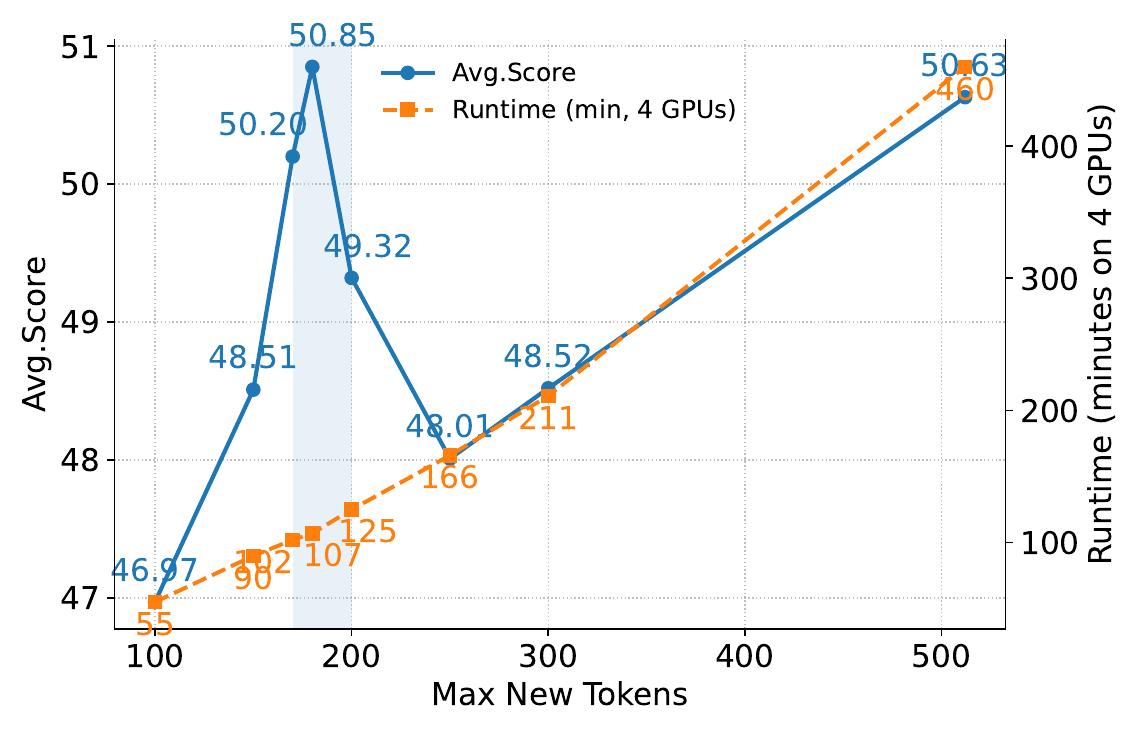}
    \caption{Trade-off of ADG scoring w.r.t.\ response sampling length on Alpaca-GPT4 dataset.}
\label{fig:eff_tradeoff}
\end{figure}

Table~\ref{tab:ablation_k} shows a clear sweet spot at $K{=}5$, which achieves the best average score and leads on most benchmarks.
When $K$ is too small (for example, $K{=}3$), the estimated dispersion and shape statistics become less stable, which can weaken ranking quality under a fixed 10K budget, even though some tasks (such as MBPP) may still benefit.
When $K$ is too large (for example, $K{\ge}7$), performance drops consistently. A plausible reason is that excessive sampling introduces more off-manifold or stylistically drifting outputs, which can distort the divergence geometry and increase noise in the selection signal, while also increasing computation.
Overall, moderate sampling provides enough diversity to reliably estimate the divergence structure while avoiding unnecessary noise and cost.

\section{Trade-off of Response Sampling Length}
\label{app:eff_tradeoff}

\paragraph{Setup.}
We ablate the maximum generation length (\texttt{max new tokens}) used when sampling multiple responses per instruction for ADG scoring.
All other settings are kept identical to our main pipeline (same pool, scoring procedure, and 10K selection budget).
We report the resulting Avg.Score together with the wall-clock runtime measured on 4 GPUs.

\paragraph{Findings.}
Figure~\ref{fig:eff_tradeoff} shows a clear trade-off between selection quality and scoring cost.
As the generation cap increases from 100 to around 170--200 tokens, Avg.Score improves substantially.
Beyond this range, the runtime grows rapidly while the quality gains become unstable, indicating diminishing returns for longer sampled responses.
Based on this trade-off, we adopt \texttt{max\_new\_tokens}=180 as a practical default in our experiments.

\begin{table*}[t]
\begin{tabular}{lccccccc}
\bottomrule[1.5pt]
\multicolumn{1}{l|}{}                        & \multicolumn{2}{c|}{\textbf{Reasoning}}            & \multicolumn{2}{c|}{\textbf{Knowledge}}                  & \multicolumn{2}{c|}{\textbf{Coding}}                    & \multirow{2}{*}{\textbf{Avg.Score}} \\ \cline{2-7}
\multicolumn{1}{l|}{}                        & \textbf{BBH} & \multicolumn{1}{c|}{\textbf{GSM8K}} & \textbf{MMLU} & \multicolumn{1}{c|}{\textbf{TruthfulQA}} & \textbf{HumanEval} & \multicolumn{1}{c|}{\textbf{MBPP}} &                                     \\ \bottomrule[1.5pt]
\multicolumn{8}{l}{\textit{Alpaca-GPT4(52K)}}                                                                                                                                                                                                                \\ \hline
\multicolumn{1}{l|}{All Data}       & 55.28        & \multicolumn{1}{c|}{36.62}          & 60.42         & \multicolumn{1}{c|}{54.84}               & 32.93              & \multicolumn{1}{c|}{42.80}          & 47.15                               \\
\multicolumn{1}{l|}{Random}         & 52.35        & \multicolumn{1}{c|}{33.28}          & 60.48         & \multicolumn{1}{c|}{53.65}               & 37.8               & \multicolumn{1}{c|}{38.80}          & 46.06                               \\
\multicolumn{1}{l|}{IFD}            & 40.35        & \multicolumn{1}{c|}{46.02}          & 63.21         & \multicolumn{1}{c|}{56.08}               & 41.46              & \multicolumn{1}{c|}{46.40}          & 48.92                               \\
\multicolumn{1}{l|}{SuperFiltering} & 53.86        & \multicolumn{1}{c|}{39.73}          & 60.13         & \multicolumn{1}{c|}{56.79}               & 40.85              & \multicolumn{1}{c|}{45.00}          & 49.39                               \\
\multicolumn{1}{l|}{SelectIT}       & 50.81        & \multicolumn{1}{c|}{33.74}          & 58.57         & \multicolumn{1}{c|}{53.89}               & 33.54              & \multicolumn{1}{c|}{41.60}          & 45.36                               \\
\multicolumn{1}{l|}{MIG}            & 51.93        & \multicolumn{1}{c|}{37.00}          & 60.08         & \multicolumn{1}{c|}{55.40}               & 35.98              & \multicolumn{1}{c|}{40.00}          & 46.73                               \\
\multicolumn{1}{l|}{SCAR}           & 53.74        & \multicolumn{1}{c|}{42.46}          & 61.43         & \multicolumn{1}{c|}{58.54}               & 33.54              & \multicolumn{1}{c|}{38.20}          & 47.99                               \\
\multicolumn{1}{l|}{Rethinking}     & 53.05        & \multicolumn{1}{c|}{36.32}          & 60.56         & \multicolumn{1}{c|}{56.04}               & 37.80              & \multicolumn{1}{c|}{38.60}          & 47.06                               \\
\multicolumn{1}{l|}{ZIP}           & 51.77        & \multicolumn{1}{c|}{37.23}          & 61.24         & \multicolumn{1}{c|}{52.33}               & 37.20              & \multicolumn{1}{c|}{40.60}          & 46.73                               \\
\multicolumn{1}{l|}{\textbf{ADG(Ours)}}      & 57.83        & \multicolumn{1}{c|}{42.84}          & 62.67         & \multicolumn{1}{c|}{57.94}               & 42.07              & \multicolumn{1}{c|}{42.20}          & \textbf{50.85}                      \\ \hline
\multicolumn{8}{l}{\textit{WizardLM(70K)}}                                                                                                                                                                                                                   \\ \hline
\multicolumn{1}{l|}{All Data}                & 56.73        & \multicolumn{1}{c|}{46.47}          & 60.50         & \multicolumn{1}{c|}{47.90}               & 43.29              & \multicolumn{1}{c|}{42.40}          & 49.55                               \\
\multicolumn{1}{l|}{Random}                  & 55.06        & \multicolumn{1}{c|}{41.70}          & 61.47         & \multicolumn{1}{c|}{52.31}               & 38.41              & \multicolumn{1}{c|}{39.20}          & 48.03                               \\
\multicolumn{1}{l|}{IFD}                     & 58.29        & \multicolumn{1}{c|}{42.38}          & 61.64         & \multicolumn{1}{c|}{53.82}               & 39.63              & \multicolumn{1}{c|}{45.40}          & 50.19                               \\
\multicolumn{1}{l|}{SuperFiltering}          & 58.58        & \multicolumn{1}{c|}{43.59}          & 61.94         & \multicolumn{1}{c|}{53.79}               & 39.02              & \multicolumn{1}{c|}{45.20}          & 50.35                               \\
\multicolumn{1}{l|}{SelectIT}                & 54.25        & \multicolumn{1}{c|}{35.78}          & 60.58         & \multicolumn{1}{c|}{46.25}               & 33.54              & \multicolumn{1}{c|}{41.80}          & 45.37                               \\
\multicolumn{1}{l|}{MIG}                     & 55.38        & \multicolumn{1}{c|}{38.51}          & 61.11         & \multicolumn{1}{c|}{50.36}               & 40.85              & \multicolumn{1}{c|}{41.80}          & 48.00                               \\
\multicolumn{1}{l|}{SCAR}                    & 52.54        & \multicolumn{1}{c|}{37.53}          & 58.84         & \multicolumn{1}{c|}{49.57}               & 29.88              & \multicolumn{1}{c|}{37.40}          & 44.29                               \\
\multicolumn{1}{l|}{Rethinking}              & 52.19        & \multicolumn{1}{c|}{37.91}          & 61.24         & \multicolumn{1}{c|}{46.82}               & 34.15              & \multicolumn{1}{c|}{39.20}          & 45.25                               \\
\multicolumn{1}{l|}{ZIP}                     & 54.37        & \multicolumn{1}{c|}{34.95}          & 61.06         & \multicolumn{1}{c|}{52.33}               & 32.93              & \multicolumn{1}{c|}{40.00}          & 45.94                               \\
\multicolumn{1}{l|}{\textbf{ADG(Ours)}}               & 57.90        & \multicolumn{1}{c|}{47.16}          & 62.55         & \multicolumn{1}{c|}{52.67}               & 44.51              & \multicolumn{1}{c|}{45.60}          & \textbf{51.73}                      \\ \hline
\multicolumn{8}{l}{\textit{CoT(100K)}}                                                                                                                                                                                                                       \\ \hline
\multicolumn{1}{l|}{All Data}                & 44.97        & \multicolumn{1}{c|}{47.38}          & 51.91         & \multicolumn{1}{c|}{43.29}               & 22.56              & \multicolumn{1}{c|}{31.20}          & 40.22                               \\
\multicolumn{1}{l|}{Random}                  & 46.40        & \multicolumn{1}{c|}{32.75}          & 53.61         & \multicolumn{1}{c|}{40.39}               & 22.56              & \multicolumn{1}{c|}{36.20}          & 38.65                               \\
\multicolumn{1}{l|}{IFD}                     & 39.30        & \multicolumn{1}{c|}{34.87}          & 60.60         & \multicolumn{1}{c|}{52.66}               & 35.37              & \multicolumn{1}{c|}{43.00}          & 44.30                               \\
\multicolumn{1}{l|}{SuperFiltering}          & 52.60        & \multicolumn{1}{c|}{37.15}          & 59.32         & \multicolumn{1}{c|}{38.75}               & 35.37              & \multicolumn{1}{c|}{41.00}          & 44.03                               \\
\multicolumn{1}{l|}{SelectIT}                & 41.31        & \multicolumn{1}{c|}{29.72}          & 53.08         & \multicolumn{1}{c|}{37.73}               & 24.39              & \multicolumn{1}{c|}{35.20}          & 36.91                               \\
\multicolumn{1}{l|}{MIG}                     & 49.49        & \multicolumn{1}{c|}{36.39}          & 57.10         & \multicolumn{1}{c|}{42.15}               & 31.10              & \multicolumn{1}{c|}{39.00}          & 42.54                               \\
\multicolumn{1}{l|}{SCAR}                    & 44.92        & \multicolumn{1}{c|}{36.62}          & 54.76         & \multicolumn{1}{c|}{42.74}               & 21.34              & \multicolumn{1}{c|}{36.80}          & 39.53                               \\
\multicolumn{1}{l|}{Rethinking}              & 47.67        & \multicolumn{1}{c|}{37.38}          & 57.38         & \multicolumn{1}{c|}{43.06}               & 26.22              & \multicolumn{1}{c|}{39.20}          & 41.82                               \\
\multicolumn{1}{l|}{ZIP}                     & 50.12        & \multicolumn{1}{c|}{37.07}          & 56.45         & \multicolumn{1}{c|}{46.28}               & 29.27              & \multicolumn{1}{c|}{38.00}          & 42.87                               \\
\multicolumn{1}{l|}{\textbf{ADG(Ours)}}              & 53.79        & \multicolumn{1}{c|}{36.31}          & 58.04         & \multicolumn{1}{c|}{43.94}               & 37.19              & \multicolumn{1}{c|}{41.40}          & \textbf{45.11}                      \\ \bottomrule[1.5pt]
\end{tabular}
\caption{Detailed results on the LLaMA3-8B-Instruct backbone.}
\label{app:llama}
\end{table*}

\begin{table*}[]
\begin{tabular}{lccccccc}
\bottomrule[1.5pt]
\multicolumn{1}{l|}{}                   & \multicolumn{2}{c|}{\textbf{Reasoning}}            & \multicolumn{2}{c|}{\textbf{Knowledge}}                  & \multicolumn{2}{c|}{\textbf{Coding}}                    & \multirow{2}{*}{\textbf{Avg.Score}} \\ \cline{2-7}
\multicolumn{1}{l|}{}                   & \textbf{BBH} & \multicolumn{1}{c|}{\textbf{GSM8K}} & \textbf{MMLU} & \multicolumn{1}{c|}{\textbf{TruthfulQA}} & \textbf{HumanEval} & \multicolumn{1}{c|}{\textbf{MBPP}} &                                     \\ \bottomrule[1.5pt]
\multicolumn{8}{l}{\textit{Alpaca-GPT4(52K)}}                                                                                                                                                                                                           \\ \hline
\multicolumn{1}{l|}{All Data}           & 60.21        & \multicolumn{1}{c|}{78.70}          & 71.79         & \multicolumn{1}{c|}{58.97}               & 53.66              & \multicolumn{1}{c|}{60.80}          & 64.02                               \\
\multicolumn{1}{l|}{Random}             & 62.17        & \multicolumn{1}{c|}{81.20}          & 73.04         & \multicolumn{1}{c|}{60.56}               & 61.59              & \multicolumn{1}{c|}{62.00}          & 66.76                               \\
\multicolumn{1}{l|}{IFD}                & 63.23        & \multicolumn{1}{c|}{83.85}          & 73.55         & \multicolumn{1}{c|}{59.20}               & 60.98              & \multicolumn{1}{c|}{62.60}          & 67.24                               \\
\multicolumn{1}{l|}{SuperFiltering}     & 63.71        & \multicolumn{1}{c|}{82.56}          & 73.29         & \multicolumn{1}{c|}{60.73}               & 61.59              & \multicolumn{1}{c|}{66.40}          & 68.05                               \\
\multicolumn{1}{l|}{SelectIT}           & 63.48        & \multicolumn{1}{c|}{79.39}          & 73.47         & \multicolumn{1}{c|}{54.18}               & 62.20              & \multicolumn{1}{c|}{60.40}          & 65.52                               \\
\multicolumn{1}{l|}{MIG}                &  65.85       & \multicolumn{1}{c|}{83.70}          &    73.55      & \multicolumn{1}{c|}{60.70}               &    59.15           & \multicolumn{1}{c|}{62.60}          &     67.59                          \\
\multicolumn{1}{l|}{SCAR}               & 66.89        & \multicolumn{1}{c|}{83.55}          & 73.47         & \multicolumn{1}{c|}{59.35}               & 56.15              & \multicolumn{1}{c|}{63.20}          & 67.10                               \\
\multicolumn{1}{l|}{Rethinking}         & 64.28        & \multicolumn{1}{c|}{81.58}          & 73.84         & \multicolumn{1}{c|}{59.73}               & 59.15              & \multicolumn{1}{c|}{62.40}          & 66.83                               \\
\multicolumn{1}{l|}{ZIP}                & 65.04        & \multicolumn{1}{c|}{80.89}          & 74.04         & \multicolumn{1}{c|}{60.79}               & 62.02              & \multicolumn{1}{c|}{64.60}          & 67.90                               \\
\multicolumn{1}{l|}{\textbf{ADG(Ours)}} & 66.64        & \multicolumn{1}{c|}{84.21}          & 73.98         & \multicolumn{1}{c|}{60.73}               & 62.80              & \multicolumn{1}{c|}{66.00}          & \textbf{69.06}                      \\ \hline
\multicolumn{8}{l}{\textit{WizardLM(70K)}}                                                                                                                                                                                                              \\ \hline
\multicolumn{1}{l|}{All Data}           & 62.88        & \multicolumn{1}{c|}{78.70}          & 72.61         & \multicolumn{1}{c|}{55.38}               & 63.41              & \multicolumn{1}{c|}{46.80}          & 63.30                               \\
\multicolumn{1}{l|}{Random}             & 63.85        & \multicolumn{1}{c|}{80.67}          & 73.82         & \multicolumn{1}{c|}{54.18}               & 58.54              & \multicolumn{1}{c|}{57.40}          & 64.74                               \\
\multicolumn{1}{l|}{IFD}                & 65.54        & \multicolumn{1}{c|}{80.89}          & 73.19         & \multicolumn{1}{c|}{58.79}               & 60.37              & \multicolumn{1}{c|}{46.00}          & 64.13                               \\
\multicolumn{1}{l|}{SuperFiltering}     & 66.99        & \multicolumn{1}{c|}{80.89}          & 73.12         & \multicolumn{1}{c|}{57.34}               & 56.10              & \multicolumn{1}{c|}{59.00}          & 65.57                               \\
\multicolumn{1}{l|}{SelectIT}           & 65.40        & \multicolumn{1}{c|}{78.17}          & 73.79         & \multicolumn{1}{c|}{54.53}               & 58.54              & \multicolumn{1}{c|}{51.80}          & 63.71                               \\
\multicolumn{1}{l|}{MIG}                & 67.01        & \multicolumn{1}{c|}{81.18}          & 73.85         & \multicolumn{1}{c|}{57.50}               & 58.15              & \multicolumn{1}{c|}{60.40}          & 66.35                               \\
\multicolumn{1}{l|}{SCAR}               & 66.26        & \multicolumn{1}{c|}{81.04}          & 72.97         & \multicolumn{1}{c|}{55.90}               & 56.71              & \multicolumn{1}{c|}{58.00}          & 65.15                               \\
\multicolumn{1}{l|}{Rethinking}         & 65.15        & \multicolumn{1}{c|}{79.83}          & 73.86         & \multicolumn{1}{c|}{55.32}               & 56.10              & \multicolumn{1}{c|}{55.60}          & 64.31                               \\
\multicolumn{1}{l|}{ZIP}                & 65.35        & \multicolumn{1}{c|}{76.88}          & 73.49         & \multicolumn{1}{c|}{56.95}               & 56.71              & \multicolumn{1}{c|}{57.80}          & 64.53                               \\
\multicolumn{1}{l|}{\textbf{ADG(Ours)}} & 67.65        & \multicolumn{1}{c|}{82.49}          & 74.28         & \multicolumn{1}{c|}{59.05}               & 60.37              & \multicolumn{1}{c|}{62.00}          & \textbf{67.64}                      \\ \hline
\multicolumn{8}{l}{\textit{CoT(100K)}}                                                                                                                                                                                                                  \\ \hline
\multicolumn{1}{l|}{All Data}           & 60.76        & \multicolumn{1}{c|}{69.67}          & 70.50         & \multicolumn{1}{c|}{47.95}               & 52.44              & \multicolumn{1}{c|}{55.20}          & 59.42                               \\
\multicolumn{1}{l|}{Random}             & 62.62        & \multicolumn{1}{c|}{70.05}          & 73.19         & \multicolumn{1}{c|}{43.04}               & 54.27              & \multicolumn{1}{c|}{57.60}          & 60.13                               \\
\multicolumn{1}{l|}{IFD}                & 61.97        & \multicolumn{1}{c|}{76.41}          & 72.85         & \multicolumn{1}{c|}{45.89}               & 56.1               & \multicolumn{1}{c|}{57.80}          & 61.84                               \\
\multicolumn{1}{l|}{SuperFiltering}     & 62.48        & \multicolumn{1}{c|}{74.37}          & 72.80         & \multicolumn{1}{c|}{47.09}               & 54.88              & \multicolumn{1}{c|}{55.40}          & 61.17                               \\
\multicolumn{1}{l|}{SelectIT}           & 62.17        & \multicolumn{1}{c|}{70.74}          & 72.52         & \multicolumn{1}{c|}{45.25}               & 54.27              & \multicolumn{1}{c|}{54.80}          & 59.96                               \\
\multicolumn{1}{l|}{MIG}                & 64.08        & \multicolumn{1}{c|}{71.72}          & 73.37         & \multicolumn{1}{c|}{45.11}               & 53.66              & \multicolumn{1}{c|}{58.40}          & 61.05                               \\
\multicolumn{1}{l|}{SCAR}               & 59.84        & \multicolumn{1}{c|}{72.25}          & 73.14         & \multicolumn{1}{c|}{41.36}               & 56.09              & \multicolumn{1}{c|}{57.00}          & 59.95                               \\
\multicolumn{1}{l|}{Rethinking}         & 63.28        & \multicolumn{1}{c|}{73.31}          & 72.66         & \multicolumn{1}{c|}{46.62}               & 52.44              & \multicolumn{1}{c|}{58.40}          & 61.12                               \\
\multicolumn{1}{l|}{ZIP}                & 60.87        & \multicolumn{1}{c|}{72.93}          & 72.45         & \multicolumn{1}{c|}{44.79}               & 57.93              & \multicolumn{1}{c|}{55.40}          & 60.73                               \\
\multicolumn{1}{l|}{\textbf{ADG(Ours)}} & 64.28        & \multicolumn{1}{c|}{76.82}          & 74.00         & \multicolumn{1}{c|}{44.80}               & 58.54              & \multicolumn{1}{c|}{57.80}          & \textbf{62.71}                      \\ \bottomrule[1.5pt]
\end{tabular}
\caption{Detailed results on the Qwen2.5-7B-Instruct backbone.}
\label{app:qwen}
\end{table*}

\begin{table*}[]
\begin{tabular}{lccccccc}
\bottomrule[1.5pt]
\multicolumn{1}{c|}{}                   & \multicolumn{2}{c|}{\textbf{Reasoning}}            & \multicolumn{2}{c}{\textbf{Knowledge}}     & \multicolumn{2}{c|}{\textbf{Coding}}                    & \multirow{2}{*}{\textbf{Avg.Score}} \\ \cline{2-7}
\multicolumn{1}{c|}{}                   & \textbf{BBH} & \multicolumn{1}{c|}{\textbf{GSM8K}} & \textbf{MMLU} & \textbf{TruthfulQA}        & \textbf{HumanEval} & \multicolumn{1}{c|}{\textbf{MBPP}} &                                     \\ \bottomrule[1.5pt]
\multicolumn{8}{l}{\textit{Alpaca-GPT4(52K)}}                                                                                                                                                                                             \\ \hline
\multicolumn{1}{l|}{All Data}           & 29.67        & \multicolumn{1}{c|}{5.38}           & 29.53         & \multicolumn{1}{c|}{44.91} & 17.07              & 21.80                               & 24.73                               \\
\multicolumn{1}{l|}{Random}             & 31.07        & \multicolumn{1}{c|}{6.75}           & 31.76         & \multicolumn{1}{c|}{38.43} & 17.07              & 22.20                               & 24.55                               \\
\multicolumn{1}{l|}{IFD}                & 31.98        & \multicolumn{1}{c|}{7.28}           & 30.52         & \multicolumn{1}{c|}{39.12} & 18.9               & 24.20                               & 25.33                               \\
\multicolumn{1}{l|}{SuperFiltering}     & 31.77        & \multicolumn{1}{c|}{6.82}           & 31.8          & \multicolumn{1}{c|}{41.92} & 17.68              & 26.20                               & 26.03                               \\
\multicolumn{1}{l|}{SelectIT}           & 30.29        & \multicolumn{1}{c|}{7.05}           & 34.23         & \multicolumn{1}{c|}{42.05} & 18.29              & 23.20                               & 25.85                               \\
\multicolumn{1}{l|}{MIG}                & 30.21        & \multicolumn{1}{c|}{4.70}           & 30.34         & \multicolumn{1}{c|}{41.49} & 17.68              & 23.60                               & 24.67                               \\
\multicolumn{1}{l|}{SCAR}               & 30.75        & \multicolumn{1}{c|}{5.69}           & 30.76         & \multicolumn{1}{c|}{42.66} & 17.68              & 22.40                               & 24.99                               \\
\multicolumn{1}{l|}{Rethinking}         & 30.02        & \multicolumn{1}{c|}{6.28}           & 31.74         & \multicolumn{1}{c|}{42.07} & 20.12              & 23.20                               & 25.57                               \\
\multicolumn{1}{l|}{ZIP}                & 30.22        & \multicolumn{1}{c|}{6.90}           & 33.62         & \multicolumn{1}{c|}{42.05} & 20.73              & 22.20                               & 25.95                               \\
\multicolumn{1}{l|}{\textbf{ADG(Ours)}} & 30.39        & \multicolumn{1}{c|}{7.28}           & 34.28         & \multicolumn{1}{c|}{43.86} & 20.62              & 24.20                               & \textbf{26.77}                      \\ \hline
\multicolumn{8}{l}{\textit{WizardLM(70K)}}                                                                                                                                                                                                \\ \hline
\multicolumn{1}{l|}{All Data}           & 29.77        & \multicolumn{1}{c|}{6.67}           & 29.75         & \multicolumn{1}{c|}{40.19} & 17.68              & 21.20                               & 24.21                               \\
\multicolumn{1}{l|}{Random}             & 30.59        & \multicolumn{1}{c|}{6.22}           & 30.96         & \multicolumn{1}{c|}{39.78} & 17.68              & 24.80                               & 25.01                               \\
\multicolumn{1}{l|}{IFD}                & 31.24        & \multicolumn{1}{c|}{5.31}           & 28.98         & \multicolumn{1}{c|}{40.31} & 18.90              & 22.40                               & 24.52                               \\
\multicolumn{1}{l|}{SuperFiltering}     & 31.25        & \multicolumn{1}{c|}{6.22}           & 27.60         & \multicolumn{1}{c|}{38.77} & 20.12              & 24.40                               & 24.73                               \\
\multicolumn{1}{l|}{SelectIT}           & 30.9         & \multicolumn{1}{c|}{6.22}           & 31.02         & \multicolumn{1}{c|}{39.49} & 19.51              & 24.80                               & 25.32                               \\
\multicolumn{1}{l|}{MIG}                & 32.27        & \multicolumn{1}{c|}{5.16}           & 32.08         & \multicolumn{1}{c|}{40.78} & 18.26              & 21.80                               & 25.06                               \\
\multicolumn{1}{l|}{SCAR}               & 31.52        & \multicolumn{1}{c|}{6.22}           & 32.29         & \multicolumn{1}{c|}{42.19} & 17.07              & 21.60                               & 25.14                               \\
\multicolumn{1}{l|}{Rethinking}         & 31.53        & \multicolumn{1}{c|}{6.52}           & 27.92         & \multicolumn{1}{c|}{40.29} & 20.12              & 22.40                               & 24.80                               \\
\multicolumn{1}{l|}{ZIP}                & 30.46        & \multicolumn{1}{c|}{6.60}           & 30.33         & \multicolumn{1}{c|}{38.42} & 19.51              & 23.40                               & 24.79                               \\
\multicolumn{1}{l|}{\textbf{ADG(Ours)}} & 32.13        & \multicolumn{1}{c|}{6.84}           & 32.78         & \multicolumn{1}{c|}{39.53} & 20.73              & 24.80                               & \textbf{26.13}                      \\ \hline
\multicolumn{8}{l}{\textit{CoT(100K)}}                                                                                                                                                                                                    \\ \hline
\multicolumn{1}{l|}{All Data}           & 25.08        & \multicolumn{1}{c|}{10.25}          & 35.07         & \multicolumn{1}{c|}{41.54} & 11.59              & 15.20                               & 23.12                               \\
\multicolumn{1}{l|}{Random}             & 26.94        & \multicolumn{1}{c|}{9.93}           & 34.21         & \multicolumn{1}{c|}{36.91} & 17.68              & 24.00                               & 24.95                               \\
\multicolumn{1}{l|}{IFD}                & 27.29        & \multicolumn{1}{c|}{9.02}           & 29.30         & \multicolumn{1}{c|}{40.10} & 19.51              & 23.20                               & 24.74                               \\
\multicolumn{1}{l|}{SuperFiltering}     & 27.80        & \multicolumn{1}{c|}{6.60}           & 34.28         & \multicolumn{1}{c|}{38.15} & 17.68              & 25.40                               & 24.99                               \\
\multicolumn{1}{l|}{SelectIT}           & 27.42        & \multicolumn{1}{c|}{7.96}           & 33.41         & \multicolumn{1}{c|}{37.79} & 18.90              & 22.60                               & 24.68                               \\
\multicolumn{1}{l|}{MIG}                & 28.50        & \multicolumn{1}{c|}{9.40}           & 30.05         & \multicolumn{1}{c|}{39.22} & 15.24              & 22.80                               & 24.20                               \\
\multicolumn{1}{l|}{SCAR}               & 28.47        & \multicolumn{1}{c|}{9.55}           & 31.91         & \multicolumn{1}{c|}{41.08} & 15.24              & 22.20                               & 24.74                               \\
\multicolumn{1}{l|}{Rethinking}         & 29.27        & \multicolumn{1}{c|}{9.86}           & 34.20         & \multicolumn{1}{c|}{39.28} & 17.07              & 23.00                               & 25.45                               \\
\multicolumn{1}{l|}{ZIP}                & 27.35        & \multicolumn{1}{c|}{11.60}          & 35.48         & \multicolumn{1}{c|}{37.20} & 15.85              & 26.20                               & 25.61                               \\
\multicolumn{1}{l|}{\textbf{ADG(Ours)}} & 28.69        & \multicolumn{1}{c|}{10.79}          & 36.10         & \multicolumn{1}{c|}{40.30} & 17.07              & 24.40                               & \textbf{26.23}                      \\ \bottomrule[1.5pt]
\end{tabular}
\caption{Detailed results under a small backbone (LLaMA3.2-1B-Instruct). We report reasoning, knowledge, and coding performance on three selection pools (Alpaca-GPT4, WizardLM, and CoT) with a fixed 10K budget. ADG consistently improves the average score over strong selection baselines, indicating that divergence-guided selection remains effective even for a lightweight model.}
\label{table:1b_model}
\end{table*}

\section{Results with a Small Backbone (LLaMA3.2-1B)}
\label{sec:appendix_llama1b}

To test whether ADG remains effective for lightweight models, we repeat the main selection experiments using LLaMA3.2-1B-Instruct as the training backbone.
We follow the same evaluation protocol as in the main paper: selecting a 10K subset from each pool (Alpaca-GPT4, WizardLM, and CoT), training with the same recipe, and evaluating on the same suite covering reasoning (BBH, GSM8K), knowledge (MMLU, TruthfulQA), and coding (HumanEval, MBPP).
All baselines are run under the same budget and pipeline for a fair comparison.

Table~\ref{table:1b_model} shows that ADG continues to deliver consistent gains under the 1B-scale backbone, achieving the best (or competitive-best) Avg.Score across the three pools.
This suggests that the proposed divergence signal is not tied to large-capacity backbones: even when the model is small and its raw generation quality is weaker, the multi-sample output geometry still provides a useful proxy for identifying training-effective instructions.
In particular, the improvements on Alpaca-GPT4 and CoT indicate that ADG helps a small model avoid spending budget on redundant or weakly informative supervision, while preserving instructions that better shape structured reasoning and coding behaviors.
Overall, these results support the robustness of ADG across model scales.

\section{Baseline Details}
\label{sec:appendix_baseline}

In this subsection, we provide concise descriptions for all baselines.

\textbf{1) MIG}: The method constructs a label graph in semantic space (labels as nodes and semantic relations as edges) and assigns each instruction example to its corresponding labels with quality-weighted contributions. It then performs selection by maximizing marginal information gain under a diminishing-returns objective, encouraging both high-quality signals and broad semantic coverage.

\textbf{2) SCAR}: The method decomposes response style into two key elements, linguistic form and instructional surprisal, and uses their consistency signals to identify instruction–response pairs that are more beneficial for efficient SFT under comparable quality. It trains a style consistency-aware response ranking model to score and rank examples, then selects a small but highly style-consistent subset for instruction tuning.

\textbf{3) IFD}: The method gives the target LLM a brief “experience” by training on a small subset that maintains semantic coverage, and then scores each candidate example with an Instruction-Following Difficulty (IFD) metric. The IFD score is computed from the loss difference between generating the answer with vs. without the instruction context, and the method selects the higher-IFD examples as training data.

\textbf{4) Superfiltering}: The method proposes a weak-to-strong data filtering pipeline where a much smaller filter model scores instruction data using difficulty-based signals. Motivated by the observed rank consistency of these difficulty scores across weak and strong models, it uses the weak model’s ranking to select a small subset for fine-tuning the larger target model at lower filtering cost.

\textbf{5) ZIP}: ZIP treats the compression ratio of a candidate set as a proxy for redundancy and greedily constructs a subset that minimizes the compression ratio to keep information dense and less repetitive. It uses a lightweight multi-stage procedure (candidate pre-filtering followed by cascaded greedy selection) to efficiently build a low-redundancy subset without relying on expensive strong-model scoring.

\textbf{6) SelectIT}: SelectIT ranks instruction-tuning examples by exploiting the target LLM’s own uncertainty as a self-reflection signal, combining token-level and sentence-level uncertainty across prompts to make scoring more robust. It then selects the highest-scoring subset for SFT, improving fine-tuning efficiency without requiring an extra external scoring model.

\textbf{7) Rethinking}: The method first partitions the instruction pool into N semantic clusters via k-means to preserve broad coverage, then allocates each cluster a quota proportional to its size. Within each cluster, it selects the examples with the largest token length, yielding a long-text-focused subset while maintaining semantic diversity through cluster-wise quotas.

\end{document}